\documentclass{ieee-ojvt}
\usepackage{cite}
\usepackage{amsmath,amssymb,amsfonts}
\usepackage{algorithmic}
\usepackage{graphicx,color}
\usepackage{textcomp}
\def\BibTeX{{\rm B\kern-.05em{\sc i\kern-.025em b}\kern-.08em
    T\kern-.1667em\lower.7ex\hbox{E}\kern-.125emX}}
\AtBeginDocument{\definecolor{ojcolor}{cmyk}{0.93,0.59,0.15,0.02}}
\def\OJlogo{\vspace{-12pt}\includegraphics[height=24pt]{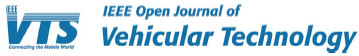}}

\usepackage{amsmath,amssymb}
\usepackage{bm}
\usepackage{listings}
\usepackage{makecell}
\usepackage{diagbox}
\usepackage{mathtools}
\usepackage{graphicx}
\usepackage{multirow}
\usepackage{rotating}
\usepackage{verbatim}
\usepackage{colortbl}
\usepackage{etoolbox}
\usepackage[normalem]{ulem}
\usepackage{booktabs}
\usepackage{lipsum}
\usepackage{stmaryrd}
\usepackage{stackengine}
\usepackage{cancel}
\usepackage{adjustbox}
\usepackage{dblfloatfix}
\usepackage{bbding}
\usepackage{wasysym}
\usepackage{algorithm}
\usepackage{float}
\usepackage[table]{xcolor}

\usepackage{glossaries}
\makeglossaries
\glsdisablehyper

\newacronym{bev}{BEV}{Bird's Eye View}
\newacronym{nerf}{NeRF}{Neural Radiance Fields}
\newacronym{lidar}{LiDAR}{Light Detection and Ranging}
\newacronym{iou}{IoU}{Intersection over Union}
\newacronym{miou}{mIoU}{mean Intersection over Union}
\newacronym{rayiou}{RayIoU}{Ray Intersection over Union}
\newacronym{mlp}{MLP}{Multi-Layer Perceptron}
\newacronym{amvae}{AM-VAE}{Air Mask Variational AutoEncoder}
\newacronym{tgp}{TGP}{Two-modal Occupancy Prediction}
\newacronym{odg}{ODG}{Occupancy Dual Gaussian}
\newacronym{vlm}{VLM}{Vision–Language Model}
\newacronym{vfm}{VFM}{Visual Foundation Model}
\newacronym{ttocc}{TT-Occ}{Test-Time-Occ}
\newacronym{fpn}{FPN}{Feature Pyramid Network}
\newacronym{dgga}{DGGA}{Dual Gaussian Graph Attention}
\newacronym{dsdga}{DSDGA}{Dynamic–Static Decoupled Gaussian Attention}
\newacronym{opus}{OPUS}{Occupancy Prediction Using a Sparse Set}
\newacronym{stwogo}{S2GO}{Streaming Sparse Gaussian Occupancy Prediction}
\newacronym{trbf}{TRBF}{Trilateral Radial Basis Function}
\newacronym{tacocc}{TACOcc}{Target-Adaptive Cross-Modal Fusion Occupancy}
\newacronym{camlid}{C+L}{camera+LiDAR}
\newacronym{camrad}{C+R}{camera+radar}
\newacronym{camlidrad}{C+L+R}{camera+LiDAR+radar}
\newacronym{camonly}{C-only}{camera-only}

\usepackage{booktabs}
\definecolor{winered}{rgb}{1, 0, 0}

\usepackage{hyperref}
\hypersetup{
    colorlinks=true,
    linkcolor=blue,
    citecolor=winered,
    filecolor=magenta,
    urlcolor=blue,
    pdfborder={0 0 0}
}

\usepackage[all]{hypcap
}% IEEE-style numbering

\definecolor{lightgray}{rgb}{0.85, 0.85, 0.85}

\usepackage{pifont}
\newcommand{\cmark}{\ding{51}}%
\newcommand{\xmark}{\ding{55}}%
\newcommand{\myangle}{90}

\usepackage{orcidlink}

\usepackage[caption=false,font=normalsize,labelfont=sf,textfont=sf]{subfig}
\usepackage{caption}

\begin{document}
\receiveddate{8 June, 2026}
\accepteddate{17 June, 2026}
\publisheddate{19 June, 2026}
\currentdate{14 July, 2026}
\doiinfo{OJVT.2026.3705622}

\title{Easy3D-Labels: Supervising Semantic Occupancy Estimation with 3D Pseudo-Labels for Automotive Perception}

\author{SEAMIE HAYES \orcidlink{0009-0008-0587-1872} \textsuperscript{1,2,3} (Graduate Student Member, IEEE), GANESH SISTU\textsuperscript{1,2}, TIM BROPHY (Member, IEEE) \orcidlink{0000-0002-1229-841X} \textsuperscript{1,2}, CIARAN EISING \orcidlink{0000-0001-8383-2635} \textsuperscript{1,2,3} (Senior Member, IEEE)}
\affil{Department of Electronic and Computer Engineering, University of Limerick, V94 T9PX Limerick, Ireland}
\affil{Data Driven Computer Engineering Research Centre, University of Limerick, V94 T9PX Limerick, Ireland}
\affil{SFI CRT Foundations in Data Science, University of Limerick, Castletroy, Co. Limerick  V94 T9PX, Ireland}
\corresp{CORRESPONDING AUTHOR: Seamie Hayes (e-mail: hayes.seamie@ul.ie).}
\authornote{This publication has emanated from research conducted with the financial support of Taighde Éireann – Research Ireland under Grant number 18/CRT/6049.}
\markboth{Easy3D-Labels: Supervising Semantic Occupancy Estimation with 3D Pseudo-Labels for Automotive Perception}{Hayes \textit{et al.}}


\begin{abstract}
In perception for automated vehicles, safety is critical not only for the driver but also for other agents in the scene, particularly vulnerable road users such as pedestrians and cyclists. Previous representation methods, such as Bird’s Eye View, collapse vertical information, leading to ambiguity in 3D object localisation and limiting accurate understanding of the environment for downstream tasks such as motion planning and scene forecasting. In contrast, semantic occupancy provides a full 3D representation of the surroundings, addressing these limitations. Furthermore, self-supervised semantic occupancy has seen increased attention in the automated vehicle domain. Unlike supervised methods that rely on manually annotated data, these approaches use 2D pseudo-labels, improving scalability by reducing the need for labour-intensive annotation. Consequently, such models employ techniques such as novel view synthesis, cross-view rendering, and depth estimation to allow for model supervision against the 2D labels. However, such approaches often incur high computational and memory costs during training, especially for novel view synthesis. To address these issues, we propose Easy3D-Labels, which are 3D pseudo-ground-truth labels generated using Grounded-SAM and Metric3Dv2, with temporal aggregation for densification, permitting supervision directly in 3D space. Easy3D-Labels can be readily integrated into existing models to provide model supervision, yielding substantial performance gains, with mIoU increasing by 45\% and RayIoU by 49\% when applied to OccNeRF on the Occ3D-nuScenes dataset. Additionally, we introduce EasyOcc, a streamlined model trained solely on these 3D pseudo-labels, avoiding the need for complex rendering strategies and achieving 15.7 mIoU on Occ3D-nuScenes. Easy3D-Labels improve scene understanding by reducing object duplication and enhancing depth estimation accuracy, as reflected by improvements in the RayIoU metric. These findings highlight the importance of foundation models, temporal information, and 3D loss formulation in self-supervised learning for comprehensive scene understanding. Our Easy3D-Labels are available open-source on \href{https://data.mendeley.com/datasets/9scymfs7xv/1}{Mendeley}.
\end{abstract}

\maketitle

\begin{figure*}[!htp]
    \centering
    \includegraphics[width=\textwidth]{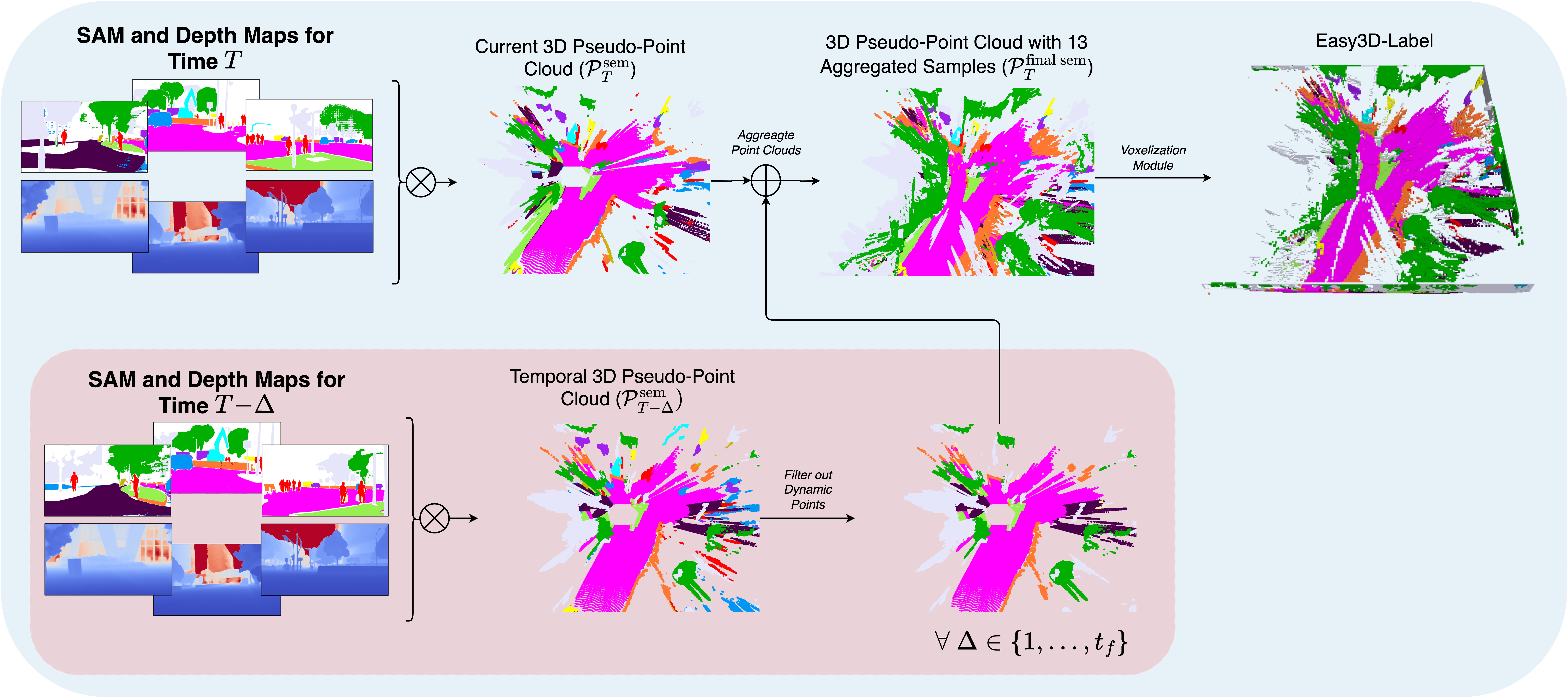}
    \caption{\textbf{Easy3D-Labels generation:} We project semantic labels into 3D using depth maps and employ temporal aggregation and object filtering for enhanced label quality.}
    \label{fig:pseudo_generation}
\end{figure*}

\section{\MakeUppercase{Introduction}}
\label{sec:introduction}
Safety in the domain of automated vehicle perception is critical to establish trust for both drivers and other road users \cite{sun2023toward}. In 2021, the global annual road traffic deaths were estimated at 1.19 million, with pedestrians accounting for 21\% of these fatalities \cite{who_2023}. It is evident that vulnerable road users account for a large proportion of road injuries and deaths. This emphasises the need for accurate and timely detection of agents in the scene to enable the vehicle to predict trajectories and make decisions in high-risk situations. 

Detection accuracy is closely tied to how the scene is represented, as restrictive representations can limit both perceptual coverage \cite{ma2024_vision_bev_survey} and the completeness of spatial reasoning \cite{mao2023_3d_bounding_survey}. Following recent progress in machine learning, many advanced approaches for automated vehicle perception have been developed \cite{xu2025_occ_survey, hayes20253d}.  One prominent approach is semantic occupancy estimation \cite{wei2023surroundocc}, whose discretised 3D representation enables more flexible scene modelling compared to earlier methods such as Bird’s Eye View and 3D bounding boxes that contain the aforementioned limitations.

Specifically, self-supervised semantic occupancy estimation is especially advantageous, largely due to its reduced reliance on manually annotated occupancy labels, offering improved scalability compared to supervised approaches \cite{huang2024selfocc}. However, these models still depend on labels in the 2D image space, using \glspl{vlm} \cite{liu2024grounding, radford2021learning, barsellotti2025talking} and \glspl{vfm} \cite{kirillov2023segment, hu2024metric3d, keetha2025mapanything} to address semantic and depth ambiguities in the absence of ground-truth 3D labels. To enable supervision in 2D, models typically use novel view synthesis techniques to render the 3D scene in image space, employing NeRF-based volume rendering \cite{mildenhall2021nerf, zhang2023occnerf} or 3D Gaussian Splatting \cite{kerbl20233dgs, gaussianocc, jiang2025gausstr}. However, relying on 2D rendering introduces high computational cost and leads to limitations such as bias towards nearby objects \cite{hayes20253d}, inaccurate detections, and object duplication \cite{sun2024gsrender}.

To address these issues, we propose a 3D labelling technique, Easy3D-Labels, which generates labels by projecting Grounded-SAM \cite{ren2024grounded} 2D semantics into 3D space using Metric3Dv2 \cite{hu2024metric3d} depth maps, illustrated in Figure \ref{fig:pseudo_generation}. This allows the model to jointly learn semantics and spatial geometry through a single pseudo-loss function. Furthermore, we aggregate temporal samples, limited to static objects, to avoid duplicating dynamic objects and to increase density. Prior work has shown the value of temporal information for improving performance \cite{tang2024sparseocc, zhang2025tt, liao2025stcocc}. The aggregation of temporal data in the supervision pipeline reduces additional computational overhead, as in other models, it is typically performed at inference time. Easy3D-Labels provides several important advantages. Firstly, the removal of the need for novel view synthesis and depth estimation during training reduces computational cost. Secondly, they enable efficient aggregation of temporal information, which is important for spatial understanding. Lastly, they support more comprehensive scene understanding in both visible and occluded regions.

Furthermore, Easy3D-Labels can be easily added to existing models as an auxiliary loss to boost performance. We explore the integration of these labels in four previous models: SelfOcc \cite{huang2024selfocc}, OccNeRF \cite{zhang2023occnerf}, GaussianOcc \cite{gaussianocc}, and GaussianFlowOcc \cite{boeder2025gaussianflowocc}. These models employ a different range of rendering techniques, scene representation, and depth supervision strategies, permitting a comprehensive analysis of the effectiveness of Easy3D-Labels. As recent advances often develop in isolation, we present a complementary method to enhance model compatibility and generalization. Additionally, we introduce EasyOcc, a streamlined model that solely uses Easy3D-Labels for loss computation, demonstrating that complex rendering techniques are not necessary for noteworthy model performance. Furthermore, our model requires no LiDAR supervision or deployment of foundation models at inference.

In summary, our main contributions are as follows:

\begin{itemize}  
    \item \textbf{Easy3D-Labels:} We introduce an approach that leverages Grounded-SAM and Metric3Dv2 to generate 3D pseudo-labels for loss computation directly in 3D space. Our 3D pseudo-labels can be effortlessly integrated into existing models via an auxiliary loss function, yielding improvements of 45\% in mIoU.
    \item \textbf{Segmentation of Dynamic Classes:} Accurate segmentation of dynamic classes is critical for safety in autonomous perception. In the case of SelfOcc, incorporating our 3D pseudo-labels improves pedestrian segmentation performance by over 600\%.
    \item \textbf{Holistic Scene Representation:} Our labels enable a more comprehensive scene representation, resulting in a 49\% increase in RayIoU for OccNeRF.
\end{itemize}

This paper is structured as follows: Section \ref{sec:lit_review} reviews related work, Section \ref{sec:methodology_models} describes the methodologies of Easy3D-Labels and the EasyOcc model, Section \ref{sec:results} presents results and ablations, and finally,  Section \ref{sec:conclusion} concludes the paper.
\section{\MakeUppercase{Literature Review}}
\label{sec:lit_review}
In Section \ref{subsec:semantic_occ}, we review semantic occupancy models, and in Section \ref{subsec:pseudolabel}, we discuss prior use of 3D pseudo-labels.

\subsection{\MakeUppercase{Semantic Occupancy Estimation}}
\label{subsec:semantic_occ}
In perception for automated vehicles, \gls{bev} methods have historically been dominant due to their simple yet effective scene representation \cite{simplebev, philion2020lift, li2024bevformer, liu2023bevfusion}. Recently, semantic occupancy estimation has gained attention, driven by benchmark datasets \cite{wei2023surroundocc, tian2023occ3d, zhu2024nucraft, tong2023scene} with accurate annotations, generated from manually labeled LiDAR data from the nuScenes automated vehicles dataset \cite{nuScenes}. This shift led to the creation of supervised occupancy estimation models \cite{tan2025geocc, ren2025rm}, with improvements from techniques such as Gaussian Splatting \cite{huang2024gaussianformer, huang2025gaussianformer2, zhao2025gaussianformer3d, song2026graphgsocc}, and multi-modal fusion \cite{hayes2025leveraging, yang2025daocc, ma2024licrocc}. 

Supervised methods rely on manually annotated LiDAR data, which is then voxelized to form ground-truth voxel annotations. Weakly supervised occupancy methods use LiDAR-derived depth and semantic labels projected into image space \cite{sun2024gsrender, pan2024renderocc, renderworld, boeder2024occflownet}, thereby avoiding the generation of occupancy ground-truth but still incurring significant data collection and annotation costs of LiDAR. In the \gls{bev} perception domain, SkyEye \cite{gosala2023skyeye} supervises with ground-truth semantic labels and pseudo-\gls{bev} labels generated via self-supervised depth estimation, although it continues to rely on manually annotated 2D labels. These limitations motivate a shift toward self-supervised alternatives, which require no manually annotated ground truth labels, improving scalability. 

In this study, we modify four self-supervised models, SelfOcc \cite{huang2024selfocc}, OccNeRF \cite{zhang2023occnerf}, GaussianOcc \cite{gaussianocc}, and GaussianFlowOcc \cite{boeder2025gaussianflowocc}, integrating them with 3D pseudo-labels for enhanced scene understanding. For the three former methods, the scene representation is in voxel space, with scene understanding occurring via the splatting of image features to 3D and a 3DCNN for refinement. The latter uses Gaussians for scene representation, with refinement occurring via induced attention \cite{lee2019set} and deformable cross-attention with image features \cite{zhu2020deformable}. 

For scene rendering, SelfOcc predicts signed distance function (SDF) based occupancy, color, and semantics via an MLP, OccNeRF via NeRF-style volume rendering \cite{mildenhall2021nerf}, and GaussianOcc and GaussianFlowOcc with Gaussian rasterization \cite{kerbl20233dgs}. For the voxel-based methods, the rendered depth supports multi-frame photometric consistency, and for GaussianFlowOcc, depth is supervised using the \gls{vfm} Metric3Dv2 \cite{hu2024metric3d}. All methods supervise semantics using 2D pseudo-labels from \glspl{vfm} \cite{ren2024grounded, zhang2023simple}. GaussianOcc further improves pose estimation through a learned 6D transformation network \cite{gaussianocc}. GaussianFlowOcc broadens the supervision space with a temporal flow module for rendering adjacent frames; however, we omit this component from training for equal comparison to the other models. Our pseudo-loss, which incorporates our Easy3D-Labels, is applied to the semantic logits voxel grid and density voxel grid, if applicable.

Other state-of-the-art methods leverage foundation models \cite{wang2025vggt, hu2024metric3d, ren2024grounded, barsellotti2025talking, keetha2025mapanything, chen2024internvl, piccinelli2024unidepth}, adopt Gaussian-based scene representations during training \cite{jiang2025gausstr, zhao2025shelfgaussian}, or rely on foundation models solely at test time \cite{zhang2025tt, zhou2025autoocc}. In addition, query-based scene representations have been explored, where occupancy and semantic labels are predicted at continuous 3D query locations, conceptually related to the continuous querying formulation used in \gls{nerf} \cite{lilja2025queryocc}.

Our proposed model, \textit{EasyOcc}, shares a similar pipeline with GaussianOcc. However, it omits several components: pose estimation, novel view synthesis, and multi-frame photometric consistency. Instead, EasyOcc solely leverages Easy3D-Labels for supervision. Despite its simplified design, EasyOcc outperforms other models that employ complex training paradigms in the mIoU and RayIoU, as detailed in Section \ref{sec:results}.
\subsection{\MakeUppercase{Pseudo-Labels}}
\label{subsec:pseudolabel}
The use of 2D pseudo-labels in self-supervised semantic occupancy models has been extensively studied, particularly through the application of \glspl{vlm} \cite{ren2024grounded, radford2021learning, barsellotti2025talking} and \glspl{vfm} \cite{piccinelli2024unidepth, hu2024metric3d}. These models have demonstrated utility across various domains, including medical applications \cite{zhang2023comprehensive, ma2024segment, wu2025medical, han2024depth}, and robotics \cite{huang2023instruct2act, chen2025vidbot}. In perception for automated vehicles, they can play a crucial role in addressing the challenge of missing ground-truth labels, particularly in resolving semantic and depth ambiguities. 

\glspl{vlm} help mitigate semantic ambiguity by leveraging both spatial and linguistic cues to produce pixel-level semantic maps. Depth ambiguity, while partially addressed using multi-frame photometric consistency, is significantly reduced through the use of metric depth \glspl{vfm} \cite{hu2024metric3d, piccinelli2024unidepth, bochkovskii2024depth}, which provide accurate pixel-level depth maps, shown to increase model performance substantially \cite{jiang2025gausstr, boeder2025gaussianflowocc}. However, to bridge the learned 3D representation to the 2D supervision space, many methods further employ NeRF-style volume rendering for supervision in semantic occupancy \cite{zhang2023occnerf, wang2024distillnerf, pan2024renderocc, huang2024selfocc, li2025semi, liu2024let} and \gls{bev} segmentation \cite{monteagudo2026rendbev}. While effective, these approaches introduce additional computational overhead, and recent work has explored Gaussian Splatting as a more efficient alternative, significantly reducing rendering cost \cite{gaussianocc, jiang2025gausstr, boeder2025gaussianflowocc}.

Following research on pseudo-labels in \gls{bev} \cite{gosala2023skyeye, monteagudo2026rendbev}, recent occupancy models have investigated the generation of 3D pseudo-labels (or shelf-labels) for model supervision \cite{zhao2025shelfgaussian, boeder2025shelfocc, li2025ago}, often leveraging large \glspl{vfm} \cite{keetha2025mapanything, oquab2023dinov2, kirillov2023segment}. These methods bring benefits such as increased performance metrics and reduced need for novel-view rendering. However, existing approaches can be limited by the spatial restriction of supervision to camera-based observations via a camera-mask \cite{boeder2025shelfocc}, requiring foundation models at inference time \cite{zhao2025shelfgaussian}, or utilizing costly LiDAR data \cite{li2025ago, zhao2025shelfgaussian}. Additionally, AutoOcc \cite{zhou2025autoocc} positions itself as a 3D pseudo-labelling pipeline. However, integrating such 3D pseudo-labels with existing models employing 2D supervision remains underexplored. 

Our proposed method, Easy3D-Labels, does not rely on LiDAR and instead combines Metric3Dv2 \cite{hu2024metric3d} for depth estimation with Grounded-SAM \cite{ren2024grounded} for semantic segmentation to generate 3D pseudo-ground-truth labels. These labels are further refined through outlier removal, occupancy thresholding, and temporal aggregation, enabling models to learn a more holistic scene representation.

\section{\MakeUppercase{Methodology}}
\label{sec:methodology_models}
This section details the Easy3D-Label generation process in Section \ref{subsec:gen_pseudo_labels}, discusses the pseudo-loss, $\mathcal{L}_{\text{Pseudo}}$, in Section \ref{subsec:pseudo_loss}, outlines model modifications in Section \ref{subsec:mod_to_existing_models}, and presents our model, EasyOcc, in Section \ref{subsec:easyocc}.

\begin{figure*}[t]
    \centering
    \includegraphics[width=\textwidth]{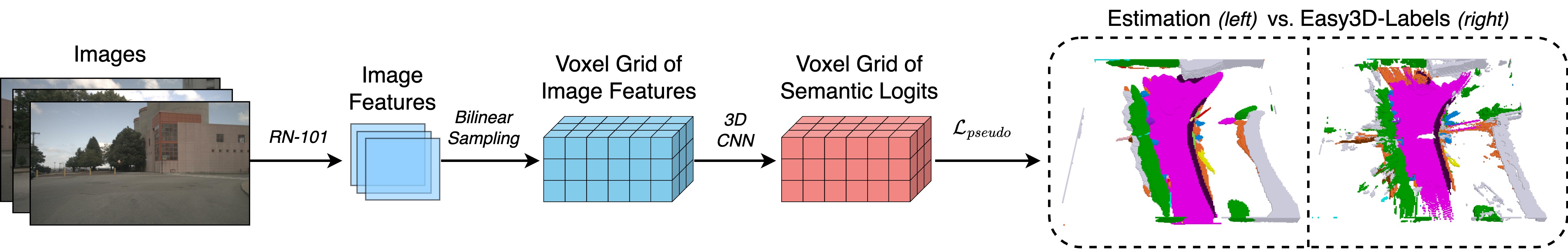}
    \caption{\textbf{EasyOcc model architecture:} Image features are extracted and then processed in voxel space by 3D convolutions prior to pseudo-loss computation against our 3D pseudo-labels, Easy3D-Labels.}
    \label{fig:easyocc}
\end{figure*}

\subsection{\MakeUppercase{Generation of Easy3D-Labels}}
\label{subsec:gen_pseudo_labels}
This section presents the key contribution of this paper: the generation of our 3D pseudo-labels.
%from semantic maps of Grounded-SAM \cite{ren2024grounded} and depth maps from Metric3Dv2 \cite{hu2024metric3d}, 
We describe the generation methods for general semantic and depth \glspl{vfm}, with generation expressed in the ego-coordinate frame of the current sample. This method enables loss computation directly in 3D voxel space, eliminating the need for view synthesis and aligning our approach with supervised training pipelines. The process is illustrated in Figure \ref{fig:pseudo_generation} and divided into three steps: semantic point cloud generation (Subsubsection \ref{subsubsec:pc_gen}), densification (Subsubsection \ref{subsubsec:pc_dense}), and voxelization (Subsubsection \ref{subsubsec:pc_voxel}).

\subsubsection{\MakeUppercase{Semantic Point Cloud Generation}}
\label{subsubsec:pc_gen}
This stage will detail the generation of a semantic point cloud for an arbitrary sample.
%Semantic maps are sourced from the OccNeRF repository \cite{zhang2023occnerf}, which are generated using Grounded-SAM, while depth maps are generated with the Giant variant of Metric3Dv2 for optimal training performance \cite{hu2024metric3d}.
For each camera, $i \in \{1, \dots, N\}$, in a sample, given the corresponding semantic map \( S_i \in \mathbb{R}^{H \times W} \), depth map \( D_i \in \mathbb{R}^{H \times W} \), camera intrinsic matrix \( K_i \in \mathbb{R}^{3 \times 3} \), and extrinsic matrix \( \mathbf{T}_{\text{camera, }i}^{\text{global}} \in \mathbb{R}^{4 \times 4} \), each arbitrary pixel \( (u, v) \in [0, W) \times [0, H)\) is projected into the homogenised global coordinates in Equation (\ref{eq:global_projection}):
\begin{equation}
\mathbf{P}_{\text{global, } i}^{(u,v)} = \mathbf{T}_{\text{camera, }i}^{\text{global}}  \begin{bmatrix}
D_i(u,v) \cdot K_i^{-1}
\begin{bmatrix}
u \\
v \\
1
\end{bmatrix} \\
1
\end{bmatrix}
\label{eq:global_projection}
\end{equation}
Following this, each projected pixel, $\mathbf{P}_{\text{global,} i}^{(u,v)}$, is dehomogenised and then decorated with its corresponding semantic pixel $S_i(u,v)$, to yield a semantic point cloud, $\mathcal{P}^{\text{sem}}_i$, seen in Equation (\ref{eq:semantic_point_cloud}), where $\mathcal{L}$ denotes the semantic label space.
\begin{equation}
\mathcal{P}^{\text{sem}}_i = \left\{ \left( \mathbf{P}_{\text{global, }i}^{(u,v)},\; S_i(u,v) \right) \right\}, \quad \mathcal{P}^{\text{sem}}_i \subset \mathbb{R}^3 \times \mathcal{L}
\label{eq:semantic_point_cloud}
\end{equation}
Finally, we aggregate the semantic point cloud for each camera, $\mathcal{P}^{\text{sem}}_i$, into a unified semantic point cloud, $\mathcal{P}^{\text{sem}}$, expressed in Equation (\ref{eq:multi_cam_aggregation}).
\begin{equation}
\mathcal{P}^{\text{sem}} = \bigcup_{i=1}^{N} \left\{ \mathcal{P}^{\text{sem}}_i \right\}
\label{eq:multi_cam_aggregation}
\end{equation}
To improve spatial accuracy, statistical outlier removal (SOR) is performed on $\mathcal{P}^{\text{sem}}$ using the Open3D library \cite{zhou2018open3d}, yielding a consolidated semantic point cloud in the global coordinate frame.

\subsubsection{\MakeUppercase{Semantic Point Cloud Densification}}
\label{subsubsec:pc_dense}
Densification of point clouds is performed using previous samples, as in real-world driving scenarios, data is naturally acquired as a continuous temporal stream. Our design reflects this by incrementally aggregating past observations to densify the scene. This streaming formulation is more practical than relying on future frames, as it enables progressive refinement of the 3D representation over time. Even in offline settings, adopting a streaming paradigm ensures that the method remains scalable to long sequences and consistent with real-world operation. This is applicable when a vehicle is deployed specifically for data acquisition or in consumer vehicles where additional data is acquired for further training (e.g., Tesla).

Following the previous step, point clouds for each sample, $\mathcal{P}_T^{\text{sem}}$, are densified using temporal semantic point clouds, $\mathcal{P}_{T{-}\Delta}^{\text{sem}}$, for all $\Delta \in \{1, \dots, t_{f}\}$ where $t_f$ denotes the number of temporal samples. First, we remove dynamic points (e.g., vehicles and pedestrians) in $\mathcal{P}_{T{-}\Delta}^{\text{sem}}$, to prevent object duplication, while retaining static points, such as the sidewalk and drivable surface. Following this, we transform the unions of $\mathcal{P}_T^{\text{sem}}$ and $\mathcal{P}_{T{-}\Delta}^{\text{sem}}$ from global coordinates to ego-vehicle coordinates of time $T$ with $\mathbf{T}_{\text{global}}^{\text{ego}} \in \mathbb{R}^{4 \times 4}$, which yields the densified semantic point cloud, $\mathcal{P}_T^{\text{final sem}}$, in Equation (\ref{eq:ego_trans}):
\begin{equation}
\label{eq:ego_trans}
\mathcal{P}_T^{\text{final sem}} = \mathbf{T}_{\text{global}}^{\text{ego}} \circ 
\left( 
\mathcal{P}_T^{\text{sem}} \cup \bigcup_{\Delta=1}^{t_f} \mathcal{P}_{T{-}\Delta}^{\text{sem}}
\right)
\end{equation}
This process ensures that the final voxelisation of $\mathcal{P}_T^{\text{final sem}}$ yields labels more closely resembling ground-truth data. The effectiveness of this step is demonstrated in the following Section \ref{subsec:qual_labels} and in the ablation study on the EasyOcc model in Section \ref{subsubsec:ablation_samples}.

\subsubsection{\MakeUppercase{Semantic Point Cloud Voxelization}}
\label{subsubsec:pc_voxel}
Once $\mathcal{P}_T^{\text{final\_sem}}$ is obtained, it is voxelized to obtain the final 3D pseudo-label according to bounds \(V_{bounds} = [X_{min}, Y_{min}, Z_{min}, X_{max}, Y_{max}, Z_{max}]\), using a voxel resolution of \(V_{res}\), expressed in the ego-frame coordinate system of the current sample. Given the high density of $\mathcal{P}_T^{\text{final\_sem}}$ due to the aggregation of many temporal samples, a voxel is considered occupied only if it contains a minimum of $\tau$ points; otherwise, it is treated as empty. This threshold helps mitigate the influence of stray points that could otherwise result in erroneous voxelization. For voxels classified as occupied, the semantic label is assigned based on the majority class among the contained points.

\begin{comment}
\begin{table}[htpb]
\centering
\begin{tabular}{c|cc}
    \hline
    \textit{\textbf{Historic Samples}}  & \textit{\textbf{Versus All (mIoU)}} & \textit{\textbf{Training Only (mIoU)}}\\
    \hline
    0 & 9.36 & 8.98 \\
    1 & 10.54 & 10.29 \\
    2 & 11.32 & 11.16 \\
    3 & 11.90 & 11.80 \\
    4 & 12.34 & 12.30 \\
    5 & 12.69 & 12.67 \\
    6 & 12.95 & 12.95 \\
    7 & 13.14 & 13.16 \\
    8 & 13.29 & 13.30 \\
    9 & 13.41 & 13.41 \\
    10 & 13.49 & 13.48 \\
    11 & 13.53 & 13.53\\
    12 & 13.56 & 13.56\\
    13 & 13.58 & 13.58 \\
    \hline
\end{tabular}
\vspace{1mm}
\caption{\textbf{Effect of the number of Historic Samples versus mIoU of Ground Truth Occ3D data:}}
\label{tab:historic_sample_miou}
\end{table}
\end{comment}

\subsection{\MakeUppercase{Pseudo Loss}}
\label{subsec:pseudo_loss}
Model supervision using Easy3D-Labels will be facilitated by the pseudo-loss function, $\mathcal{L}_{\text{Pseudo}}$.
%Now that we have described the 3D pseudo-label generation process, we shall outline the loss function for using the labels during training. The loss function, pseudo-loss, will be used in the chosen modified models and EasyOcc. 
This loss consists of two distinct terms, as shown in (\ref{eq:pseudo_loss}), where $\lambda$ is a constant initialized at the start of training (an ablation on the value of $\lambda$ is discussed in Table \ref{tab:lambda_ablation}). The formulation is adapted from GaussianOcc, where it was initially used as an optional component for training with ground-truth labels \cite{gaussianocc}.
\begin{equation}
\label{eq:pseudo_loss}
    \mathcal{L}_{\text{Pseudo}} = \mathcal{L}_{\text{CE}} + \lambda \mathcal{L}_{\text{Geometry}}
\end{equation}
As shown in Equation (\ref{eq:geometry_loss}), geometry loss is composed of three separate losses: geometric scale loss, semantic scale loss, and Lovász softmax loss \cite{berman2018lovasz}.
\begin{equation}
\label{eq:geometry_loss}
    \mathcal{L}_{\text{Geometry}} = \mathcal{L}_{\text{geom\_scal}} + \mathcal{L}_{\text{sem\_scal}} + \mathcal{L}_{\text{Lovász}}
\end{equation}
All losses, including cross-entropy, are standard in semantic occupancy estimation, as they effectively penalize misclassifications and support class re-weighting to address dataset imbalance.
\subsection{\uppercase{Modifications to Existing Models}}
\label{subsec:mod_to_existing_models}
Incorporating $\mathcal{L}_{\text{Pseudo}}$ into the four selected architectures requires considering how the loss function interacts with existing losses and also specific implementation details, as outlined below.

\textbf{SelfOcc:} As described in Section \ref{subsec:semantic_occ}, SelfOcc predicts semantic occupancy using a two-step process: binary occupancy prediction ($occ$), followed by a semantic voxel prediction ($sem$), which together produce the final output. This structure introduces two key considerations: (1) a voxel may be classified as occupied by $occ$ but empty by $sem$, resulting in it being considered unoccupied, and (2) a voxel may be classified as unoccupied by $occ$ but occupied by $sem$, leading to it remaining unoccupied. 

Through preliminary testing, we observe that excluding $occ$ from pseudo-loss computation and from the final scene representation improved performance in both IoU and mIoU metrics. This is perhaps explained by the considerations discussed above. Hence, the final scene representation is the semantic voxel, $sem$. The final loss function is defined in Equation (\ref{eq:selfocc_loss}). The additional losses present aid in SDF stability, multi-frame photometric consistency, RGB rendering, and 2D semantic loss.
\begin{equation}
\label{eq:selfocc_loss}
\mathcal{L}_{\text{SelfOcc}} = \mathcal{L}_{\text{regularisation}} + \mathcal{L}_{\text{reprojection}} + \mathcal{L}_{\text{rgb}} + \mathcal{L}_{\text{sem}} + \mathcal{L}_{\text{Pseudo}}
\end{equation}

%To adapt 3D pseudo-labels for SelfOcc, we follow the EasyOcc design with minor modifications. In addition to the pseudo-loss, $\mathcal{L}_{\text{Pseudo}}$, we introduce an auxiliary loss for occupancy, $\mathcal{L}_{\text{occ}}$, computed as the binary cross-entropy (BCE) loss between the negated logits of $occ$ and the binary occupancy pseudo-labels, as defined in Equation (\ref{eq:occ_loss}). Here, $x_i$ represents the predicted logits, and $y_i \in \{0, 1\}$ denotes the ground-truth occupancy labels. The logits $x_i$ are negated because occupancy is derived from the signed distance function (SDF), where values less than zero indicate occupancy, contrasting with the usual convention where positive values represent occupancy.
%
%\begin{equation}
%\label{eq:occ_loss}
%\mathcal{L}_{\text{occ}} = \frac{1}{N} \sum_{i=1}^N \left[ \log(1 + e^{x_i}) - x_i y_i \right],
%\end{equation}

\textbf{OccNeRF:} Here, we implement pseudo-loss alongside additional loss components to form the final loss function, as shown in Equation (\ref{eq:occnerf_loss}). The pseudo-loss serves as a complement to the three existing losses in the original OccNeRF model: $\mathcal{L}_{\text{regularisation}}$, $\mathcal{L}_{\text{reprojection}}$, $\mathcal{L}_{\text{sem}}$. These losses regulate voxel occupancy stability, multi-frame photometric consistency, and 2D semantic loss, respectively.
\begin{equation}
\label{eq:occnerf_loss}
\mathcal{L}_{\text{OccNeRF}} = \mathcal{L}_{\text{regularisation}} + \mathcal{L}_{\text{reprojection}} + \mathcal{L}_{\text{sem}} + \mathcal{L}_{\text{Pseudo}}
\end{equation}

\textbf{GaussianOcc:} Given the similarity between GaussianOcc and OccNeRF, our pseudo-loss implementation follows the same approach in GaussianOcc, with one key difference: we omit $\mathcal{L}_{\text{sem}}$ due to sporadic NaN gradients in the convolutional layers of the image encoder during training. The cause of this issue remains unknown. The resulting loss function is defined in Equation (\ref{eq:gaussianocc_loss}).
\begin{equation}
\label{eq:gaussianocc_loss}
\mathcal{L}_{\text{GaussianOcc}} =  \mathcal{L}_{\text{regularisation}} +\mathcal{L}_{\text{reprojection}} + \mathcal{L}_{\text{Pseudo}}
\end{equation}

\textbf{GaussianFlowOcc:} This method operates similarly to SelfOcc: binary occupancy prediction ($occ$), followed by a semantic voxel prediction ($sem$), which are used to get the final prediction. Since the semantic voxel prediction in GaussianFlowOcc does not include an explicit empty-class logit, it does not encode free-space occupancy directly. Therefore, losses that require a full semantic occupancy distribution, such as the geometric and semantic scale losses, are not applied to $sem$. Instead, we supervise $sem$ using Cross-Entropy over occupied semantic classes, while supervising the separate occupancy branch, $occ$, using Binary Cross-Entropy.
\begin{equation}
\label{eq:gaussianflowocc_loss}
\mathcal{L}_{\text{GaussianFlowOcc}} =  \mathcal{L}_{\text{sem}} +\mathcal{L}_{\text{depth}} + \mathcal{L}_{\text{CE}} +
\mathcal{L}_{\text{BCE}}
\end{equation}
\subsection{\MakeUppercase{EasyOcc}}
\label{subsec:easyocc}
Integrating $\mathcal{L}_{\text{Pseudo}}$ in EasyOcc is straightforward, as the framework relies exclusively on this signal for learning. The continuous and dense scene representation enables effective learning in conjunction with our 3D pseudo-labels.

The model is a simplified variant of GaussianOcc, with its architectural flow illustrated in Figure \ref{fig:easyocc}. Multi-view camera images are processed through a ResNet-101 image encoder to extract high-level features, which have shown robust performance across \gls{bev} models \cite{simplebev} and semantic occupancy prediction frameworks \cite{gaussianocc}. Parameter-free bilinear sampling projects these features into 3D space, which are then passed through a 3D CNN to enhance spatial reasoning and produce semantic logits. EasyOcc eliminates the need for depth estimation, multi-frame consistency, and novel view synthesis, thus reducing training complexity and duration.

\begin{table*}[t]
\centering
\caption{\textbf{Model configurations:} Rep. denotes scene representation; SAM - Grounded-SAM; Pho. - photometric consistency; M3D - Metric3Dv2. Rendering time is per sample for training-time generation of semantics, depth, or features; training time is per epoch. Parentheses (+) indicate pseudo-loss overhead; * OccNeRF and ** GaussianOcc parameter counts include training-only NeRF rendering and pose estimation modules, respectively. Training time is expressed in GPU hours.
%(-) indicates the data was not listed in the paper or repository. 
%TT-OccCamera and AutoOcc are not listed as they require no training.
}
\begin{tabular}{l|cc|cc|ccc|cc}
\toprule
\multirow{2}{*}{\textbf{Method}} 
& \multirow{2}{*}{\textbf{Rep.}} 
& \multirow{2}{*}{\textbf{Epochs}}
& \multicolumn{2}{c|}{\textbf{Input}} 
& \multicolumn{3}{c|}{\textbf{Supervision}} 
& \multicolumn{2}{c}{\textbf{Time}} \\

& 
& 
& \textbf{Backbone} & \textbf{Size}
& \textbf{Space} & \textbf{Sem.} & \textbf{Depth}
& \textbf{Render} & \textbf{Train} \\
\midrule
SelfOcc \cite{huang2024selfocc} & Voxel & 24 & RN-50 & 800$\times$384 & 2D & OpenSeed & Pho. & 32ms & 8hr 40m \textit{(+80m)} \\
OccNeRF \cite{zhang2023occnerf} & Voxel & 24 & RN-101 & 672$\times$336 & 2D & SAM & Pho. & 1061ms & 20hr 40m \textit{(+120m)} \\
GaussianOcc \cite{gaussianocc} & Voxel & 24 & RN-101 & 640$\times$384 & 2D & SAM & Pho. & 23ms & 6hr 0m \textit{(+40m)} \\
GaussianFlowOcc \cite{boeder2025gaussianflowocc} & Gaussian & 24 & RN-50 & 704$\times$256 & 2D & SAM & Metric3Dv2 & 2ms & 3hr 20m \textit{(+80m)} \\
\midrule
EasyOcc (Ours) & Voxel & 24 & RN-101 & 640$\times$384 & 3D & SAM & Metric3Dv2 & 0ms & 5hr 40m \\
\bottomrule
\end{tabular}
\label{tab:sota_models_config}
\end{table*}

\section{\uppercase{Results}}
\label{sec:results}
In this section, we present the main results. Sections \ref{subsec:dataset_metrics}, \ref{subsec:implementation}, and \ref{subsec:model_config} describe the dataset, metrics, implementation details, and configurations. We analyze occupancy metrics for mIoU and RayIoU in Sections \ref{subsec:sota_results_cam} and \ref{subsec:rayiou}, respectively, and the quantitative results of Easy3D-Labels in Section \ref{subsec:qual_labels}. We then provide an ablation study of EasyOcc in Section \ref{subsec:ablation_easyocc} and qualitative analysis in Section \ref{subsec:qual_results}.

\subsection{\uppercase{Dataset and Evaluation Metrics}}
\label{subsec:dataset_metrics}
We use two datasets, Occ3D-nuScenes \cite{tian2023occ3d} and OpenOccv2 \cite{tong2023scene}, and three evaluation metrics: \gls{iou}, \gls{miou}, and RayIoU. On Occ3D-nuScenes, all models are evaluated using \gls{iou}, \gls{miou}, and RayIoU. On OpenOccv2, models are evaluated only using RayIoU because the dataset does not provide a camera mask, preventing fair \gls{miou} evaluation for self-supervised models. However, OpenOccv2 contains denser labels, enabling a fairer evaluation under the RayIoU metric. Both datasets are derived from nuScenes and contain 600 training scenes and 150 validation scenes. The voxel space is bounded by \([-40\text{m}, -40\text{m}, -1\text{m}, 40\text{m}, 40\text{m}, 5.4\text{m}]\), with a voxel size of \(0.4\text{m}\).

\gls{iou}, defined in Equation (\ref{eq:iou}), reflects the model’s ability to capture overall spatial structure through occupancy. The \gls{miou} metric, shown in Equation (\ref{eq:miou}), computes the average \gls{iou} across all semantic classes, excluding the empty class.
\begin{equation}
\label{eq:iou}
\text{IoU} = \frac{TP}{TP + FP + FN}
\end{equation}
\begin{equation}
\label{eq:miou}
\text{mIoU} = \frac{1}{C} \sum_{\substack{c=1}}^{C} \frac{TP_c}{TP_c + FP_c + FN_c}
\end{equation}
\begin{center}
\text{TP: True Positive,\quad FP: False Positive,\quad FN: False Negative}
\end{center}
RayIoU is defined similarly to \gls{miou}, but instead of being a voxel-wise metric, it is a ray metric. Introduced in SparseOcc \cite{tang2024sparseocc}, RayIoU aims to resolve the harsh penalisation of incorrect depth estimations and the overcompensation of overprediction of voxels for inflating the mIoU score. A ray is cast from the LiDAR sensor position in both the ground-truth and predicted voxel grids, and it is labelled correct if both ray depths are within a threshold and are the same class. The equation is:
\begin{equation}
\label{eq:rayiou}
\text{RayIoU} = \frac{1}{C} \sum_{\substack{c=1}}^{C} \frac{TP_c}{TP_c + FP_c + FN_c}
\end{equation}
\subsection{\MakeUppercase{Implementation Details}}
\label{subsec:implementation}
For the 3D pseudo-label generation process, semantic maps are sourced from the OccNeRF repository \cite{zhang2023occnerf}, which are generated using Grounded-SAM, while depth maps are generated with the Giant variant of Metric3Dv2 for optimal training performance \cite{hu2024metric3d}, with 6 total cameras from nuScenes. We set $t_f=13$ historical frames due to memory constraints, and set the occupancy threshold $\tau=10$. We use the Occ3D-nuScenes \cite{tian2023occ3d} dataset and hence, the semantic label space is $\mathcal{L} = \{0, 1, \dots, 17\}$, voxel bounds \(V_{bounds} =  [-40\text{m}, -40\text{m}, -1\text{m}, 40\text{m}, 40\text{m}, 5.4\text{m}]\), with a voxel resolution \(V_{res} = 0.4\text{m}\) expressed in the ego-frame coordinate system of the current sample.

In the pseudo-loss $\mathcal{L}_{pseudo}$, we set $\lambda=0.1$. Self-supervised models that utilize LiDAR during training or inference are excluded from our comparison \cite{li2025ago, zhang2025tt, sze2025minkocc, zheng2024veon} as our method is a camera-only pipeline, consistent with the models used for comparison in this study. Training and inference are performed on four NVIDIA A100-SXM4-40GB GPUs. Models are trained for 24 epochs with a batch size of 1.

\subsection{\MakeUppercase{Model Configurations \& Inference Time}}
\label{subsec:model_config}
In this section, we discuss preliminaries regarding model configurations, label generation, and model inference time.

\subsubsection{\MakeUppercase{Model Configurations}}
In Table \ref{tab:sota_models_config}, we compare all model configurations. Rendering time and depth estimation add overhead, with Gaussian Splatting being the most efficient due to its rasterization-based rendering \cite{kerbl20233dgs}. EasyOcc avoids these rendering methods during training, resulting in reduced training time. While incorporating pseudo-loss increases epoch training time, the impact varies by model, with OccNeRF experiencing the largest increase (+120m) due to its overall slower training.

\subsubsection{\MakeUppercase{Easy3D-Label Generation Time}}
We report the time required to generate Grounded-SAM, Metric3Dv2, and Easy3D-Labels in Table \ref{tab:label_generation}. As both Grounded-SAM and Metric3Dv2 are also required by other methods, they are not considered additional overhead for our label generation. Time is expressed in GPU hours and can be reduced through parallelisation and batching. Compared with the model training times reported in Table \ref{tab:sota_models_config}, our label generation time is of the same order as a single training epoch, indicating that label generation integrates efficiently with the overall training pipeline.

%We also identify areas for further improvement, particularly in outlier removal, which could be accelerated through voxel thresholding, similar to the approach used during voxelization.

\begin{table}[htpb]
\centering
\caption{\textbf{Easy3D-Labels Generation Time.} Easy3D-Label generation time is measured from the end of generating the pseudo-labels to generating the voxel grid for the entire training set in GPU hours.}
\begin{tabular}{c|c}
    \toprule
    \textbf{Process} & \textbf{Generation Time}\\
    \midrule
    Grounded-SAM & 4d 5h\\
    Metric3Dv2 & 20h \\
    \midrule
    Easy3D-Labels & 10h \\
    \bottomrule
\end{tabular}
\label{tab:label_generation}
\end{table}

\subsubsection{\MakeUppercase{EasyOcc Inference Time}}
In Table \ref{tab:infer_time_per_step}, we report the inference time, in milliseconds, for each component of the EasyOcc model. Grid Sampling denotes the downsampling of the contracted-coordinate voxel grid to match the dimensions of the Occ3D ground-truth labels, following the approach introduced in OccNeRF \cite{zhang2023occnerf}. The contracted coordinate system enables the unbounded camera-visible scene to be represented within a finite voxel grid, particularly motivated by the absence of metric depth labels. We retain this formulation to allow direct comparison with prior works \cite{zhang2023occnerf, gaussianocc}. Additionally, preliminary experiments further indicated that using the contracted coordinate system, rather than modeling the scene directly in the Occ3D output space, led to improved performance.

\begin{table}[h]
\centering
\caption{\textbf{EasyOcc inference time breakdown.}}
\begin{tabular}{c|c}
    \toprule
    \textbf{Process} & \textbf{Execution Time}\\
    \midrule
    Image Encoding & 27ms \\
    Bilinear Sampling & 6ms \\
    3D CNN & 7ms \\
    Grid Sampling & 145ms \\
    \midrule
    Total & 185ms \\
    \bottomrule
\end{tabular}
\label{tab:infer_time_per_step}
\end{table}
\subsection{\MakeUppercase{Main Results: mIoU}}
\label{subsec:sota_results_cam}
In this section, we evaluate the performance of the four selected baseline models and EasyOcc in Table \ref{tab:miou_table}. We compare the models across four key evaluation categories: inference time (FPS), Intersection over Union (IoU), mean IoU (mIoU), and class-wise IoU for each semantic category. We provide further models for reference in the table.

% mIoU

\definecolor{others}{rgb}{0, 0, 0}
\definecolor{barrier}{rgb}{1, 0.47058824, 0.19607843}
\definecolor{bicycle}{rgb}{1, 0.75294118, 0.79607843}
\definecolor{bus}{rgb}{1, 1, 0.0}
\definecolor{car}{rgb}{0.0, 0.58823529, 0.96078431}
\definecolor{construction}{rgb}{0, 1, 1}
\definecolor{motorcycle}{rgb}{0.78431372549 , 0.70588235294, 0}
\definecolor{pedestrian}{rgb}{1, 0, 0}
\definecolor{cone}{rgb}{1, 0.94117647, 0.58823529}
\definecolor{trailer}{rgb}{0.52941176, 0.23529412, 0}
\definecolor{truck}{rgb}{0.62745098, 0.1254902, 0.94117647}
\definecolor{driveable}{rgb}{1, 0, 1}
\definecolor{flat}{rgb}{0.54509804,0.5372549,0.5372549}
\definecolor{sidewalk}{rgb}{0.29411765,0,0.29411765}
\definecolor{terrain}{rgb}{0.58823529,0.94117647,0.31372549}
\definecolor{manmade}{rgb}{0.90196078,0.90196078,0.98039216}
\definecolor{vegetation}{rgb}{0,0.68627451,0}

\begin{table*}[ht]
  \caption{\textbf{State-of-the-art of camera-only self-supervised occupancy comparison} on the Occ3D-nuScenes \cite{tian2023occ3d} dataset. * OccNeRF uses 2D semantic loss, while GaussianOcc does not. ** GaussianFlowOcc trained without the temporal flow module. FPS denotes frames per second, indicating processing time per sample. IoU and mIoU refer to Intersection over Union and mean Intersection over Union. Grey rows indicate SOTA methods for reference. mIoU does not include \textit{others} and \textit{other\_flat} classes. The best result for each model (compared to its variant with our labels) is highlighted in \textbf{bold}.}
\centering
  \setlength{\tabcolsep}{4pt}
  \begin{adjustbox}{width=\textwidth}
    \begin{tabular}{
        l
        |>{\columncolor{green!3}}c
        | >{\columncolor{blue!3}}c
          >{\columncolor{blue!3}}c
        |*{15}{c}
    }
    \toprule
\textbf{Method} & \textbf{FPS} & \textbf{IoU} & \textbf{mIoU} & \rotatebox{\myangle}{\textcolor{barrier}{$\blacksquare$} barrier} & \rotatebox{\myangle}{\textcolor{bicycle}{$\blacksquare$} bicycle} & \rotatebox{\myangle}{\textcolor{bus}{$\blacksquare$} bus} & \rotatebox{\myangle}{\textcolor{car}{$\blacksquare$} car} & \rotatebox{\myangle}{\textcolor{construction}{$\blacksquare$} const. veh.} & \rotatebox{\myangle}{\textcolor{motorcycle}{$\blacksquare$} motorcycle} & \rotatebox{\myangle}{\textcolor{pedestrian}{$\blacksquare$} pedestrian} & \rotatebox{\myangle}{\textcolor{cone}{$\blacksquare$} traffic cone} & \rotatebox{\myangle}{\textcolor{trailer}{$\blacksquare$} trailer} & \rotatebox{\myangle}{\textcolor{truck}{$\blacksquare$} truck} & \rotatebox{\myangle}{\textcolor{driveable}{$\blacksquare$} drive. surf.} & \rotatebox{\myangle}{\textcolor{sidewalk}{$\blacksquare$} sidewalk} & \rotatebox{\myangle}{\textcolor{terrain}{$\blacksquare$} terrain} & \rotatebox{\myangle}{\textcolor{manmade}{$\blacksquare$} manmade} & \rotatebox{\myangle}{\textcolor{vegetation}{$\blacksquare$} vegetation} \\
    \midrule 
    \rowcolor{black!4}
    DistillNeRF \cite{wang2024distillnerf} & 2.8 & 29.1 & 10.1 & 1.4 & 2.1 & 10.2 & 10.1 & 2.6 & 2.0 & 5.5 & 4.6 & 1.4 & 7.9 & 43.0 & 16.9 & 15.0 & 14.1 & 15.1 \\
    \rowcolor{black!4}
    GaussTR \cite{jiang2025gausstr} & 0.3 & 44.5 & 13.8 & 6.5 & 8.5 & 21.8 & 24.3 & 6.3 & 15.5 & 7.9 & 1.9 & 6.1 & 17.2 & 37.0 & 17.2 & 7.2 & 21.2 & 10.0 \\
    \rowcolor{black!4}
    TT-Occ \cite{zhang2025tt} & 0.7 & - & 16.7 & 21.5 & 10.5 & 10.7 & 14.7 & 11.9 & 12.3 & 9.7 & 12.2 & 4.4 & 7.9 & 48.3 & 23.7 & 28.3 & 14.1 & 20.2 \\
    \rowcolor{black!4}
    GaussianFlowOcc \cite{boeder2025gaussianflowocc} & 9.6 & 46.9 & 17.1 & 6.8 & 9.7 & 19.0 & 17.2 & 4.2 & 11.8 & 9.3 & 10.3 & 1.8 & 12.3 & 61.0 & 31.2 & 34.8 & 14.7 & 12.4 \\
    \rowcolor{black!4}
    AutoOcc \cite{zhou2025autoocc} & - & 83.0 & 20.9 & 12.7 & 10.5 & 7.8 & 20.4 & 5.8 & 17.6 & 18.5 & 24.3 & 4.2 & 12.9 & 55.5 & 24.2 & 27.1 & 35.6 & 36.6 \\
        \rowcolor{black!4}
    QueryOcc \cite{lilja2025queryocc} & 11.6 & 55.0 & 21.3 & 7.3 & 6.8 & 26.5 & 20.9 & 4.8 & 10.9 & 15.0 & 13.2 & 3.7 & 17.3 & 69.2 & 34.5 & 38.4 & 25.2 & 25.7 \\
        \rowcolor{black!4}
    ShelfOcc \cite{boeder2025shelfocc} & -- & 56.1 & 22.9 & 14.0 & 11.4 & 25.3 & 25.8 & 7.3 & 16.6 & 12.9 & 13.4 & 5.4 & 17.2 & 68.0 & 34.7 & 42.7 & 25.6 & 22.9 \\
\midrule
SelfOcc \cite{huang2024selfocc} & 7.4 & \textbf{44.1} & 10.3 & 0.2 & 0.5 & 6.7 & 10.4 & 0.0 & 0.1 & 2.1 & 0.0 & 0.0 & 7.7 & \textbf{56.1} & \textbf{26.9} & \textbf{25.7} & \textbf{13.4} & 4.6 \\
+ Ours & 7.4 & 34.5 & \textbf{14.4} & \textbf{1.7} & \textbf{5.4} & \textbf{14.5} & \textbf{22.2} & \textbf{2.6} & \textbf{6.4} & \textbf{15.4} & \textbf{8.9} & \textbf{1.0} & \textbf{12.8} & 55.0 & 26.8 & 21.6 & 11.3 & \textbf{9.3} \\
\midrule
OccNeRF \cite{zhang2023occnerf} & 5.4 & \textbf{46.4} & 11.0 & 0.7 & 1.8 & 6.6 & 6.6 & \textbf{3.7} & 0.3 & 2.9 & 3.2 & \textbf{2.9} & 6.6 & 52.8 & 24.0 & \textbf{25.0} & \textbf{18.6} & 9.7 \\
+ Ours$^*$ & 5.4 & 38.5 & \textbf{16.0} & \textbf{1.9} & \textbf{8.2} & \textbf{16.7} & \textbf{22.1} & 1.0 & \textbf{7.7} & \textbf{14.7} & \textbf{12.8} & 1.0 & \textbf{13.8} & \textbf{55.9} & \textbf{28.0} & 22.7 & 15.8 & \textbf{17.2} \\
\midrule
GaussianOcc \cite{gaussianocc} & 5.4 & \textbf{42.9} & 11.3 & \textbf{1.8} & 5.8 & 14.6 & 13.6 & 1.3 & 2.8 & 8.0 & 9.8 & 0.6 & 9.6 & 44.6 & 20.1 & 17.6 & 8.6 & 10.3 \\
+ Ours$^*$ & 5.4 & 38.8 & \textbf{15.7} & 1.7 & \textbf{5.9} & \textbf{16.2} & \textbf{22.3} & \textbf{2.3} & \textbf{8.4} & \textbf{15.7} & \textbf{10.2} & \textbf{1.0} & \textbf{13.4} & \textbf{55.3} & \textbf{27.4} & \textbf{23.4} & \textbf{16.0} & \textbf{17.0}\\
\midrule
%GaussianFlowOcc \cite{boeder2025gaussianflowocc} & 9.6 & 40.5 & 16.7 & 6.7 & 9.2 & 17.5 & 16.5 & 4.0 & 11.9 & 8.8 & 9.7 & 0.6 & 12.3 & 49.3 & 30.8 & 34.6 & 17.4 & 20.8 
GaussianFlowOcc$^{**}$ \cite{boeder2025gaussianflowocc} & 9.6 & 35.9 & 12.0 & \textbf{5.5} & 1.9 & 9.3 & 9.4 & 0.3 & 1.9 & 3.0 & 3.0 & 1.5 & 7.3 & 57.8 & 22.3 & \textbf{28.3} & 13.6 & 14.7
\\
+ Ours & 9.6 & \textbf{39.6} & \textbf{16.2} & 1.4 & \textbf{7.3} & \textbf{17.6} & \textbf{20.6} & \textbf{4.0} & \textbf{7.0} & \textbf{8.2} & \textbf{7.2} & 1.5 & \textbf{14.3} & \textbf{60.3} & \textbf{29.3} & 26.3 & \textbf{16.8} & \textbf{20.8}

\\
\midrule
EasyOcc (Ours) & 5.4 & 38.9 & 15.7 & 1.9 & 6.7 & 15.1 & 21.7 & 2.7 & 8.1 & 15.3 & 11.1 & 1.4 & 12.8 & 55.8 & 27.9 & 22.1 & 16.1 & 17.3 \\
%EasyGaussOcc (Ours) & 9.6 & 37.5 & 16.2 & 1.5 & 5.7 & 16.2 & 23.0 & 1.9 & 6.4 & 10.4 & 6.5 & 0.7 & 14.0 & 63.5 & 31.3 & 29.2 & 14.3 & 18.6 \\
    % OLD NO BETTER LOVASZ
    %SelfOcc \cite{selfocc} + Ours & \textbf{6.7} & 34.55 & 13.31 & 1.67 & 5.99 & 13.48 & 21.27 & 2.48 & 6.30 & \textbf{15.43} & 10.06 & 0.75 & 12.02 & 54.68 & 26.78 & 21.78 & 11.34 & 8.87 \\
    %OccNeRF \cite{occnerf} + Ours * & \underline{5.4} & 38.35 & \textbf{13.91} & 1.81 & \underline{8.21} & \underline{15.89} & \underline{21.83} & 0.68 & 7.93 & 14.16 & \textbf{13.42} & 0.52 & \underline{14.03} & 54.76 & \underline{27.79} & 22.58 & 15.90 & 16.92 \\
    %GaussianOcc \cite{gaussianocc} + Ours * & \underline{5.4} & 38.76 & 13.66 & 1.74 & 6.17 & 15.49 & 20.10 & 1.89 & \underline{8.40} & 15.19 & 10.28 & 0.96 & 12.88 & \underline{55.82} & 27.68 & 22.74 & 15.93 & \underline{17.01} \\
    %EasyOcc (Ours) RN-152 & 5.1 & - & 13.78 & 1.80 & 7.80 & 15.02 & 20.04 & 1.98 & 8.69 & 15.39 & 11.03 & 1.36 & 12.34 & 55.12 & 27.51 & 22.38 & 16.13 & 17.75 \\
    %EasyOcc (Ours) RN-101 & 5.4 & - & 13.48 & 1.80 & 6.12 & 14.04 & 20.14 & 2.16 & 8.72 & 14.81 & 10.32 & 0.71 & 12.43 & 55.36 & 27.58 & 22.44 & 15.69 & 16.89 \\
    %EasyOcc (Ours) RN-18 & 6.0 & - & 12.84 & 1.66 & 4.75 & 12.40 & 20.26 & 2.70 & 6.88 & 13.76 & 9.74 & 1.02 & 10.91 & 54.81 & 26.81 & 21.67 & 14.76 & 16.15 \\
    \bottomrule
  \end{tabular}
  \end{adjustbox}
  \label{tab:miou_table}
\end{table*}

\subsubsection{\MakeUppercase{Inference Time}}
GaussianFlowOcc achieves the highest inference speed at 9.6 FPS, attributed to its sparse Gaussian representation and usage of induced attention, which significantly reduces computational overhead. SelfOcc ranks second with 7.4 FPS, benefiting from the lack of a contracted coordinate system. The inclusion of pseudo-loss has no effect on inference time, as it influences only the training phase. EasyOcc matches the inference speed of both OccNeRF and GaussianOcc, all of which operate at 5.4 FPS. Both GaussTR and TT-Occ exhibit poor FPS due to the deployment of large foundation models at inference time.

\subsubsection{\MakeUppercase{Intersection over Union}}
AutoOcc secures the highest performance with an IoU of 83.01, significantly surpassing all other models. Methods employing 3D supervision rank second and third \cite{boeder2025shelfocc, lilja2025queryocc}. Notably, the addition of pseudo-loss led to a marked decline in IoU except for GaussianFlowOcc; for example, it reduces SelfOcc's score from 44.1 to 34.5, a 22\% drop. This performance drop is attributed to object duplication, as the model lacks the ability to reason about occluded regions. Consequently, it tends to predict occupancy beyond visible surfaces, resulting in an overly dense scene representation, which is further evidenced in the evaluation of RayIoU in Section \ref{subsec:rayiou}. This is visualized in the qualitative analysis presented in Section \ref{subsec:qual_results}. The lack of a drop for GaussianFlowOcc can be attributed to the sparsity of Gaussians, leading to less object duplication.

\begin{comment}
\begin{table}[htpb]
\centering
\begin{tabular}{c|cc}
    \hline
    \textbf{Model} & \textbf{IoU} & \textbf{mIoU} \\
    \hline
    GaussianOcc \cite{gaussianocc} & 5.53 & 5.91 \\
    GaussianOcc \cite{gaussianocc} + Ours & \textbf{17.01} & \textbf{7.71} \\
    GaussTR \cite{gausstr} & 11.58 & 4.47 \\
    \hline
\end{tabular}
\vspace{1mm}
\caption{\textbf{Evaluation Without Camera Mask:} GaussianOcc and GaussianOcc with pseudo-loss, and GaussTR evaluated without the camera voxel mask. The best performer in each category is highlighted in \textbf{bold}.}
\label{tab:iou_cam_mask}
\end{table}

The results clearly show that our 3D pseudo-labels enable the model to develop a more holistic scene understanding, significantly outperforming both GaussianOcc and GaussTR in the mIoU and IoU metrics. This supports the conclusion that, when not constrained by the camera mask, the IoU metric alone is not a reliable indicator of overall scene understanding, whereas mIoU offers a more balanced measure. Interestingly, GaussTR performs worse than GaussianOcc in this setting, despite outperforming it with the camera mask applied. This raises a pertinent question: should self-supervised models be evaluated solely on visible voxels, or should they be held to the same standard as supervised models and assessed across the entire voxel space?
\end{comment}

\subsubsection{\MakeUppercase{Mean Intersection over Union}}
A notable performance gap exists between the original voxel-based models \cite{huang2024selfocc, zhang2023occnerf, gaussianocc} and those employing a Gaussian scene representation and 3D supervision. ShelfOcc \cite{boeder2025shelfocc} leads due to high-quality 3D supervision integration into the SOTA supervised model architecture STCOcc \cite{liao2025stcocc}.

With the sole usage of Easy3D-Labels, EasyOcc reaches an mIoU of 15.7, surpassing the Gaussian-based method GaussTR. The employment of pseudo-loss allows OccNeRF to achieve a 15\% improvement over GaussTR and a 45\% gain compared to the original OccNeRF model. Similar performance boosts are observed for both SelfOcc and GaussianOcc. Notably, SelfOcc gains the ability to predict previously unsupported classes, such as construction vehicles, thanks to semantic supervision from Grounded-SAM. Deployment of Easy3D-Labels into the Gaussian-based GaussianFlowOcc sees great performance gains, comparable to the voxel-based methods, showing transferability of our labels to other scene representation methods. Furthermore, despite the lack of usage of rendering for loss, EasyOcc achieves results on par with the other four models, which use Easy3D-Labels, displaying the strength of our labels.

% SOTA NO MASK
%\input{tables/sota_table_no_mask}

\begin{table*}[htp]
    \caption{\textbf{RayIoU evaluated on the Occ3D-nuScenes \cite{tian2023occ3d} dataset:} RayIoU means do not include \textit{others} and \textit{other\_flat} classes. * OccNeRF implements 2D semantic loss, whereas GaussianOcc does not. ** GaussianFlowOcc trained without the temporal flow module. The best-performer is highlighted in \textbf{bold}, for each model compared to the same model trained with our labels integrated.}
     %\andreic{Are mem. and fps identical for SuperQuadricOcc and SQ-Nerf identical?}
  \centering
  \setlength{\tabcolsep}{4pt}
  \begin{adjustbox}{width=\textwidth}
  \begin{tabular}{
    @{}l
    | >{\columncolor{blue!3}}c
    |  >{\columncolor{blue!3}}c
        >{\columncolor{blue!3}}c
        >{\columncolor{blue!3}}c
    |*{15}{c}
    @{}
}
    \toprule
\textbf{Method} 
& \rotatebox{\myangle}{\textbf{RayIoU}}
& \rotatebox{\myangle}{\textbf{RayIoU@1m}}
& \rotatebox{\myangle}{\textbf{RayIoU@2m}}
& \rotatebox{\myangle}{\textbf{RayIoU@4m}}
& \rotatebox{\myangle}{\textcolor{barrier}{$\blacksquare$} barrier} & \rotatebox{\myangle}{\textcolor{bicycle}{$\blacksquare$} bicycle} & \rotatebox{\myangle}{\textcolor{bus}{$\blacksquare$} bus} & \rotatebox{\myangle}{\textcolor{car}{$\blacksquare$} car} & \rotatebox{\myangle}{\textcolor{construction}{$\blacksquare$} const. veh.} & \rotatebox{\myangle}{\textcolor{motorcycle}{$\blacksquare$} motorcycle} & \rotatebox{\myangle}{\textcolor{pedestrian}{$\blacksquare$} pedestrian} & \rotatebox{\myangle}{\textcolor{cone}{$\blacksquare$} traffic cone} & \rotatebox{\myangle}{\textcolor{trailer}{$\blacksquare$} trailer} & \rotatebox{\myangle}{\textcolor{truck}{$\blacksquare$} truck} & \rotatebox{\myangle}{\textcolor{driveable}{$\blacksquare$} drive. surf.} & \rotatebox{\myangle}{\textcolor{sidewalk}{$\blacksquare$} sidewalk} & \rotatebox{\myangle}{\textcolor{terrain}{$\blacksquare$} terrain} & \rotatebox{\myangle}{\textcolor{manmade}{$\blacksquare$} manmade} & \rotatebox{\myangle}{\textcolor{vegetation}{$\blacksquare$} vegetation} \\
    \midrule 
    SelfOcc \cite{huang2024selfocc} & 10.9 & 7.6 & 10.9 & 14.1 & 0.9 & 1.0 & 15.5 & 15.7 & 0.0 & 0.8 & 5.0 & 0.0 & 0.0 & 19.7 & 48.4 & 13.6 & \textbf{17.5} & \textbf{16.4} & 8.7 \\
+ Ours & \textbf{16.4} & \textbf{12.3} & \textbf{16.6} & \textbf{20.3} & \textbf{3.1} & \textbf{4.7} & \textbf{23.5} & \textbf{28.5} & \textbf{3.5} & \textbf{5.4} & \textbf{22.4} & \textbf{13.4} & \textbf{1.2} & \textbf{26.1} & \textbf{49.5} & \textbf{17.5} & 14.7 & 15.0 & \textbf{17.8} \\
\midrule
OccNeRF \cite{zhang2023occnerf} & 11.8 & 7.8 & 11.6 & 16.0 & 2.4 & 2.1 & \textbf{28.3} & 21.0 & \textbf{4.6} & 1.0 & 5.8 & 7.5 & 0.6 & 16.5 & 39.8 & 12.5 & 11.4 & 13.1 & 10.9 \\
+ Ours$^*$  & \textbf{17.6} & \textbf{13.2} & \textbf{17.7} & \textbf{21.7} & \textbf{3.1} & \textbf{6.2} & 26.6 & \textbf{28.9} & 3.1 & \textbf{6.2} & \textbf{22.6} & \textbf{15.6} & \textbf{2.0} & \textbf{28.3} & \textbf{50.7} & \textbf{18.4} & \textbf{15.6} & \textbf{15.7} & \textbf{20.4} \\
\midrule
GaussianOcc \cite{gaussianocc} & 13.4 & 9.8 & 13.5 & 16.9 & 2.6 & \textbf{6.6} & \textbf{23.6} & 18.0 & 2.0 & 2.5 & 11.9 & \textbf{12.9} & 0.7 & 22.5 & 42.3 & 16.3 & 14.5 & 11.7 & 13.3 \\
+ Ours$^*$ & \textbf{16.2} & \textbf{12.1} & \textbf{16.3} & \textbf{20.2} & \textbf{2.8} & 3.5 & 22.8 & \textbf{28.3} & \textbf{3.6} & \textbf{6.1} & \textbf{21.0} & 12.0 & \textbf{0.8} & \textbf{24.5} & \textbf{49.5} & \textbf{17.8} & \textbf{14.6} & \textbf{15.3} & \textbf{20.2} \\
\midrule
GaussianFlowOcc$^{**}$ \cite{boeder2025gaussianflowocc} & 11.8 & 7.4 & 11.3 & 16.6 & \textbf{11.0} & 1.7 & 25.1 & 21.8 & 0.5 & 2.6 & 8.1 & 4.6 & 0.2 & 24.2 & 31.7 & 7.8 & 8.8 & 14.5 & 13.8 \\
+ Ours & \textbf{15.9} & \textbf{10.8} & \textbf{15.7} & \textbf{21.0} & 3.5 & \textbf{4.5} & \textbf{30.1} & \textbf{29.5} & \textbf{4.5} & \textbf{5.7} & \textbf{13.0} & \textbf{9.6} & \textbf{1.1} & \textbf{28.8} & \textbf{44.7} & \textbf{12.1} & \textbf{11.0} & \textbf{18.2} & \textbf{21.5} \\
\midrule
EasyOcc (Ours) & 16.5 & 12.3 & 16.7 & 20.6 & 3.0 & 4.6 & 21.9 & 27.7 & 4.5 & 6.4 & 20.7 & 12.7 & 2.5 & 26.3 & 50.3 & 17.6 & 14.4 & 15.3 & 20.2 \\
    \bottomrule
  \end{tabular}
  \end{adjustbox}
  \label{tab:rayiou_class_occ3d}
\end{table*}

\begin{table*}[htp]
    \caption{\textbf{RayIoU evaluated on the OpenOccv2 \cite{tong2023scene} dataset:} RayIoU means do not include \textit{others} and \textit{other\_flat} classes. * OccNeRF implements 2D semantic loss, whereas GaussianOcc does not. ** GaussianFlowOcc trained without the temporal flow module. The best-performer is highlighted in \textbf{bold}, for each model compared to the same model trained with our labels integrated.}
     %\andreic{Are mem. and fps identical for SuperQuadricOcc and SQ-Nerf identical?}
  \centering
  \setlength{\tabcolsep}{4pt}
  \begin{adjustbox}{width=\textwidth}
  \begin{tabular}{
    @{}l
    | >{\columncolor{blue!3}}c
    |  >{\columncolor{blue!3}}c
        >{\columncolor{blue!3}}c
        >{\columncolor{blue!3}}c
    |*{15}{c}
    @{}
}
    \toprule
\textbf{Method} 
& \rotatebox{\myangle}{\textbf{RayIoU}}
& \rotatebox{\myangle}{\textbf{RayIoU@1m}}
& \rotatebox{\myangle}{\textbf{RayIoU@2m}}
& \rotatebox{\myangle}{\textbf{RayIoU@4m}}
& \rotatebox{\myangle}{\textcolor{barrier}{$\blacksquare$} barrier} & \rotatebox{\myangle}{\textcolor{bicycle}{$\blacksquare$} bicycle} & \rotatebox{\myangle}{\textcolor{bus}{$\blacksquare$} bus} & \rotatebox{\myangle}{\textcolor{car}{$\blacksquare$} car} & \rotatebox{\myangle}{\textcolor{construction}{$\blacksquare$} const. veh.} & \rotatebox{\myangle}{\textcolor{motorcycle}{$\blacksquare$} motorcycle} & \rotatebox{\myangle}{\textcolor{pedestrian}{$\blacksquare$} pedestrian} & \rotatebox{\myangle}{\textcolor{cone}{$\blacksquare$} traffic cone} & \rotatebox{\myangle}{\textcolor{trailer}{$\blacksquare$} trailer} & \rotatebox{\myangle}{\textcolor{truck}{$\blacksquare$} truck} & \rotatebox{\myangle}{\textcolor{driveable}{$\blacksquare$} drive. surf.} & \rotatebox{\myangle}{\textcolor{sidewalk}{$\blacksquare$} sidewalk} & \rotatebox{\myangle}{\textcolor{terrain}{$\blacksquare$} terrain} & \rotatebox{\myangle}{\textcolor{manmade}{$\blacksquare$} manmade} & \rotatebox{\myangle}{\textcolor{vegetation}{$\blacksquare$} vegetation} \\
    \midrule
    %SelfOcc \cite{huang2024selfocc} & 9.1 & 5.8 & 9.1 & 12.3 & 15.3 & 20.1 & 0.0 & 14.4 & 0.0 & 0.9 & 0.8 & 4.8 & 0.0 & 0.6 & 39.0 & 11.8 & 13.6 & 15.7 & 7.3\\
    %+ Ours & 15.1 & 11.2 & 15.3 & 18.7 & 28.5 & 26.6 & 1.3 & 23.2 & 3.5 & 4.3 & 5.1 & 22.7 & 12.4 & 2.9 & 45.4 & 16.3 & 18.0 & 16.2 & 19.2\\
    %\midrule
    %OccNeRF \cite{zhang2023occnerf} & 11.4 & 7.9 & 11.3 & 15.0 & 21.6 & 17.3 & 0.6 & 29.4 & 4.3 & 1.8 & 0.9 & 4.9 & 6.2 & 2.1 & 41.8 & 13.8 & 13.8 & 12.7 & 9.6\\
    %+ Ours  & 15.9 & 12.0 & 16.1 & 19.6 & 28.3 & 28.9 & 2.0 & 25.8 & 3.4 & 5.3 & 5.9 & 22.5 & 14.3 & 2.9 & 46.5 & 17.4 & 18.5 & 15.4 & 21.0\\
    %\midrule
    %GaussianOcc \cite{gaussianocc} & 11.7 & 8.5 & 11.8 & 14.8 & 17.7 & 23.3 & 0.6 & 22.8 & 2.0 & 5.8 & 2.2 & 11.0 & 11.1 & 2.4 & 37.8 & 14.9 & 13.7 & 11.1 & 12.2\\
    %+ Ours & 14.9 & 11.1 & 15.1 & 18.5 & 28.0 & 24.9 & 1.1 & 22.4 & 4.0 & 3.1 & 5.8 & 20.8 & 10.7 & 2.6 & 47.0 & 17.3 & 18.1 & 15.2 & 20.7 \\
    %\midrule
    %EasyOcc (Ours) & 15.1 & 11.3 & 15.4 & 18.7 & 27.3 & 26.5 & 3.1 & 21.6 & 5.0 & 4.1 & 6.0 & 20.6 & 11.4 & 2.9 & 46.7 & 17.1 & 18.2 & 15.0 & 20.7\\
SelfOcc \cite{huang2024selfocc} 
& 9.7 & 6.2 & 9.6 & 13.1 & 0.8 & 0.9 & 14.0 & 14.7 & 0.0 & 0.8 & 4.8 & 0.0 & 0.0 & 18.9 & 39.4 & 12.1 & 14.2 & \textbf{16.2} & 8.1 \\
+ Ours 
& \textbf{16.1} & \textbf{12.0} & \textbf{16.4} & \textbf{20.0} & \textbf{2.8} & \textbf{4.1} & \textbf{22.5} & \textbf{27.5} & \textbf{3.6} & \textbf{4.9} & \textbf{21.8} & \textbf{12.1} & \textbf{1.5} & \textbf{25.4} & \textbf{45.9} & \textbf{16.2} & \textbf{18.1} & 15.9 & \textbf{19.1} \\

\midrule
OccNeRF \cite{zhang2023occnerf} 
& 12.1 & 8.4 & 12.1 & 16.0 & 2.1 & 1.7 & \textbf{29.0} & 21.0 & \textbf{4.6} & 0.9 & 5.1 & 6.2 & 0.6 & 16.8 & 42.1 & 14.2 & 14.4 & 12.9 & 10.5 \\
+ Ours$^*$  
& \textbf{17.0} & \textbf{12.8} & \textbf{17.2} & \textbf{20.9} & \textbf{2.8} & \textbf{5.0} & 25.1 & \textbf{27.4} & 3.3 & \textbf{5.6} & \textbf{21.7} & \textbf{13.8} & \textbf{2.0} & \textbf{27.7} & \textbf{47.1} & \textbf{17.4} & \textbf{18.7} & \textbf{15.8} & \textbf{21.1} \\

\midrule
GaussianOcc \cite{gaussianocc} 
& 12.5 & 9.1 & 12.6 & 15.8 & 2.4 & \textbf{5.4} & \textbf{21.9} & 17.4 & 2.1 & 2.2 & 10.8 & \textbf{10.7} & 0.6 & 22.2 & 38.3 & 15.1 & 14.0 & 11.3 & 12.8 \\
+ Ours$^*$ 
& \textbf{15.9} & \textbf{11.9} & \textbf{16.1} & \textbf{19.7} & \textbf{2.6} & 2.9 & 21.7 & \textbf{27.0} & \textbf{3.9} & \textbf{5.7} & \textbf{20.1} & 10.4 & \textbf{1.1} & \textbf{23.6} & \textbf{47.5} & \textbf{17.3} & \textbf{18.4} & \textbf{15.4} & \textbf{20.8} \\

\midrule
GaussianFlowOcc$^{**}$ \cite{boeder2025gaussianflowocc} 
& 12.6 & 8.3 & 12.4 & 17.3 & \textbf{10.8} & 1.6 & 25.3 & 22.3 & 0.5 & 2.6 & 8.0 & 4.5 & 0.2 & 25.2 & 36.4 & 10.5 & 13.0 & 14.9 & 14.0 \\
+ Ours 
& \textbf{16.5} & \textbf{11.6} & \textbf{16.5} & \textbf{21.3} & 3.2 & \textbf{4.1} & \textbf{28.8} & \textbf{28.3} & \textbf{4.4} & \textbf{5.5} & \textbf{12.8} & \textbf{8.9} & \textbf{1.3} & \textbf{28.8} & \textbf{47.7} & \textbf{15.5} & \textbf{16.3} & \textbf{18.9} & \textbf{22.4} \\

\midrule
EasyOcc (Ours) & 16.2 & 12.1 & 16.4 & 20.0 & 2.8 & 3.8 & 20.8 & 26.4 & 4.7 & 5.8 & 19.9 & 11.0 & 3.1 & 25.3 & 47.2 & 17.1 & 18.4 & 15.4 & 20.8 \\
    %EasyGaussOcc (Ours) & 13.0 & 9.3 & 13.1 & 16.8 & 3.1 & 1.8 & 22.9 & 24.3 & 2.1 & 4.9 & 12.0 & 7.8 & 0.2 & 21.4 & 47.3 & 14.9 & 14.4 & 14.1 & 18.9 \\
    \bottomrule
  \end{tabular}
  \end{adjustbox}
  \label{tab:rayiou_class_openocc}
\end{table*}

\subsubsection{\MakeUppercase{IoU per Semantic Class}}
For single-class IoU, increasing supervision space by harnessing temporal information greatly increases segmentation capabilities, namely for static classes such as \textit{sidewalk} and \textit{terrain}. Interestingly,  GaussTR achieves notable performance for dynamic objects such as \textit{bus} and \textit{car}, despite not using temporal information or flow modelling, owing to the use of foundation models at inference time. Evidently, models not using metric depth supervision \cite{huang2024selfocc, zhang2023occnerf, gaussianocc} struggle with many classes, namely vulnerable classes such as \textit{pedestrian}, only achieving noteworthy performance for large common classes such as \textit{sidewalk}.

With the integration of Easy3D-Labels, results increase significantly across nearly all categories. For GaussianOcc, IoU increases in 14 out of 15 classes, with the \textit{barrier} class as the sole exception. Similar trends are seen in the other four models. We observe that performance generally improves for dynamic classes, while some static classes, particularly large-area categories such as \textit{manmade}, do not show the same gains, potentially due to overprediction aiding their segmentation in models without our labels. Notably, SelfOcc demonstrates a dramatic 624\% increase in IoU for the \textit{pedestrian} class, emphasizing the importance of 3D labels for accurately detecting vulnerable object categories. Interestingly, GaussianFlowOcc does not exhibit the same increase for the \textit{pedestrian} class, which warrants further analysis, as other small dynamic classes (\textit{bicycle}, \textit{motorcycle}) see similar gains as seen in the other models. EasyOcc exhibits strong detection capabilities, achieving performance comparable to that of the modified models enhanced with pseudo-loss. 
\subsection{\MakeUppercase{Main Results: RayIoU}}
\label{subsec:rayiou}
In this section, we analyse models on the RayIoU metric across varying distance thresholds, along with class-wise RayIoU in Tables \ref{tab:rayiou_class_occ3d} and \ref{tab:rayiou_class_openocc}. As discussed previously, this metric penalises overprediction, which can otherwise inflate mIoU without accurately reflecting overall scene understanding. Additionally, evaluation across two datasets enables a more robust and consistent analysis.

Across both tables, training with Easy3D-Labels leads to substantial improvements in RayIoU across all distance thresholds. For instance, OccNeRF improves by 49\% on Occ3D and by 40\% on OpenOccv2. Consistent with the mIoU results, OccNeRF achieves the strongest overall performance, reaching RayIoU scores of 17.6 and 17.0 on the two datasets, respectively. Although OccNeRF and GaussianFlowOcc have the same initial RayIoU on Occ3D-nuScenes, OccNeRF benefits more from training with our pseudo-labels, outperforming GaussianFlowOcc by 1.7 RayIoU. This may be because OccNeRF’s voxel-based scene representation is better aligned with the structure of the pseudo-labels.
Evaluating across different distance thresholds also highlights improved depth accuracy, with RayIoU@1m for SelfOcc increasing by over 93\% on the OpenOccv2 dataset. Furthermore, these results emphasise the overprediction issue present in the original models, where IoU scores were artificially inflated, which RayIoU penalizes.

For class-wise RayIoU, Easy3D-Labels again enable more accurate detection, consistent with trends observed in the mIoU results, particularly for dynamic objects. Vulnerable classes, such as \textit{pedestrian}, show significant improvements; for example, OccNeRF on Occ3D achieves an increase of over 300\%, largely due to the accuracy provided by Metric3Dv2 \cite{hu2024metric3d}. Similar to the IoU analysis, the \textit{pedestrian} class does not see such a large increase in RayIoU for both datasets in GaussianFlowOcc. Additionally, SelfOcc on OpenOccv2 shows improvements across all classes, with notable gains for \textit{car}, \textit{motorcycle}, and \textit{pedestrian}. Our model, EasyOcc, demonstrates strong and competitive performance compared to SelfOcc trained with Easy3D-Labels, despite not relying on rendering-based losses.
\subsection{\MakeUppercase{Ablation Study: Easy3D-Labels}}
\label{subsec:qual_labels}
In this section, we compare our Easy3D-Labels with the ground-truth labels from Occ3D-nuScenes \cite{nuScenes} and discuss label generation time.

\subsubsection{\MakeUppercase{Temporal Samples}}
\begin{figure}[htpb]
    \centering
    \includegraphics[width=.45\textwidth]{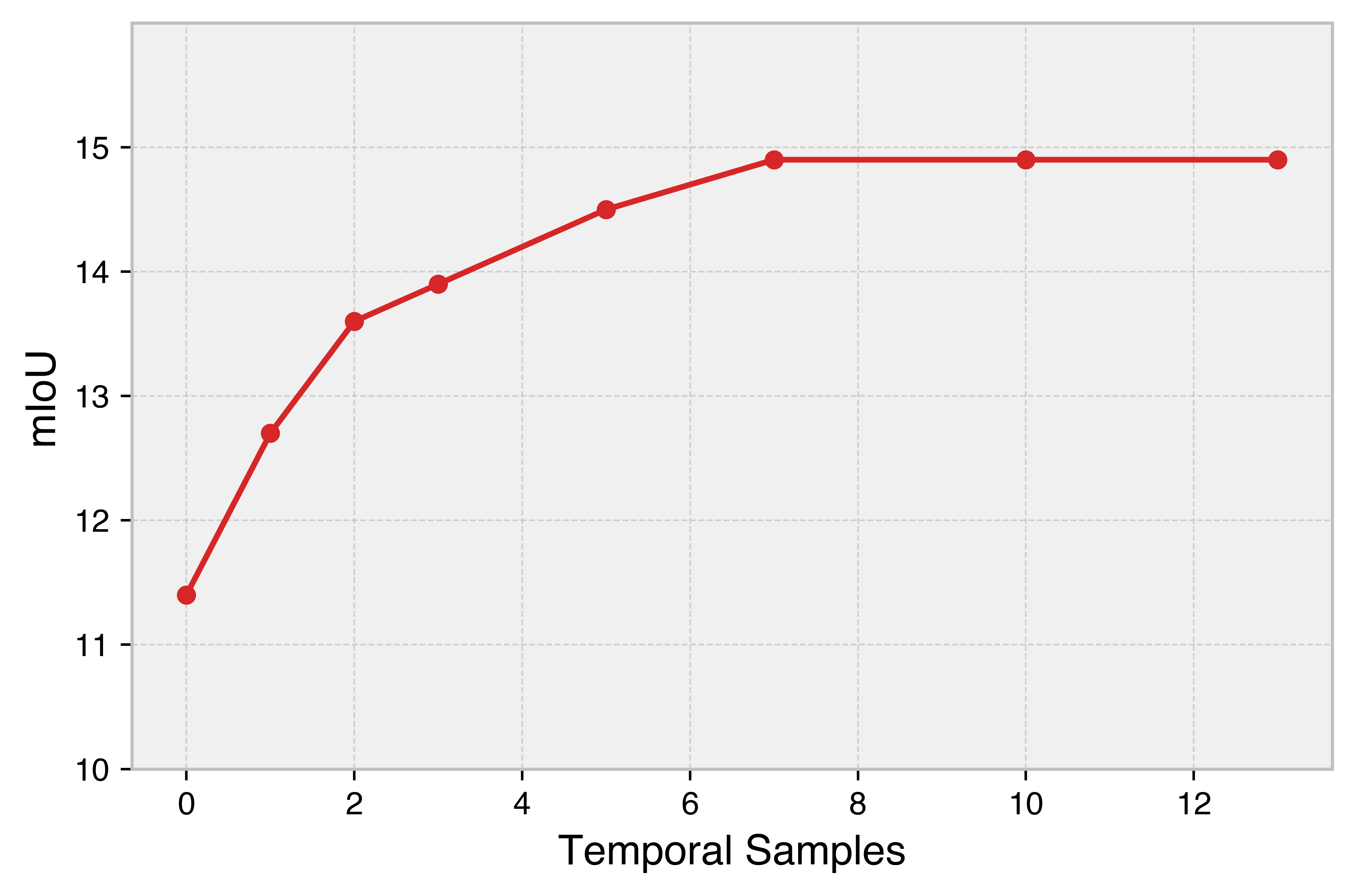}
    \caption{\textbf{Temporal sample aggregation:} Easy3D-Labels compared to Occ3D-nuScenes \cite{tian2023occ3d} labels for various numbers of aggregated samples.}    
    \label{fig:temporal_aggregation}
\end{figure}

In Figure \ref{fig:temporal_aggregation}, we compare Easy3D-Labels using varying numbers of aggregated temporal samples against the Occ3D-nuScenes ground-truth labels. The result follows a logarithmic trend, indicating saturation, where aggregating more temporal samples provides diminishing returns in mIoU. The optimal number of temporal samples is found to be $t_f=13$, at which point we achieve the highest mIoU score of 15.4, which is further evidenced in Figure \ref{fig:iou_ablation}. These findings show that incorporating temporal samples improves the similarity of the pseudo-labels to the Occ3D ground truth. We cap $t_f=13$ due to memory constraints.

\subsubsection{\MakeUppercase{Threshold Value}}
\begin{figure}[htpb]
    \centering
    \includegraphics[width=.45\textwidth]{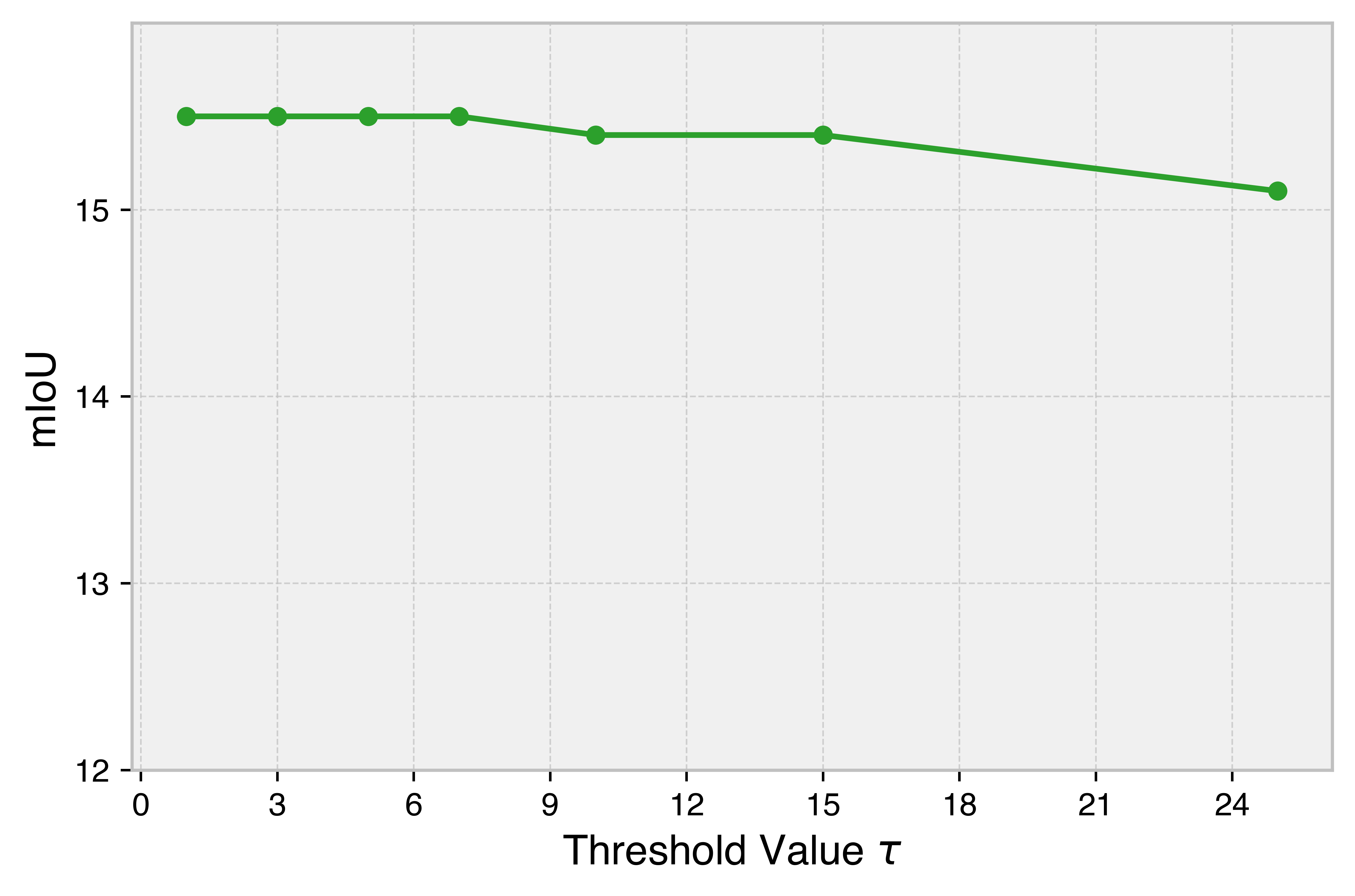}
    \caption{\textbf{Occupancy threshold:} Easy3D-Labels compared to Occ3D-nuScenes \cite{tian2023occ3d} labels for various threshold values $\tau$ in the generation process.}    
    \label{fig:threshold_analysis}
\end{figure}

In Figure \ref{fig:threshold_analysis}, we compare our labels generated using different occupancy threshold values $\tau$, ranging from 1 to 25. The highest performance is observed at a threshold of 3, indicating that even noisy points contribute useful information. For our experiments, we selected a threshold of $\tau=10$ to balance slightly faster generation time with comparable accuracy.

\subsection{\MakeUppercase{Ablation Study: EasyOcc}}
\label{subsec:ablation_easyocc}
In this section, we present a series of ablation studies on the EasyOcc model to assess the impact of key design choices. We ablate the image encoder, temporal samples, and pseudo-loss, among others. Models are trained for 12 epochs for computational efficiency.

%First, in Subsubsection \ref{subsubsec:ablation_image}, we examine the effect of varying the image encoder and input image resolution. This is followed by an analysis of temporal sample aggregation in Subsubsection \ref{subsubsec:ablation_samples}. Next, in Subsubsection \ref{subsubsec:ablation_lambda}, we evaluate the sensitivity of pseudo-loss to changes in the weighting constant $\lambda$. In Subsubsection \ref{subsubsec:ablation_losses}, we investigate the effect of selectively removing components of the pseudo-loss. Finally, in Subsubsection \ref{subsubsec:ablation_lovasz}, we explore the performance of Lovász-Softmax loss in more detail. All evaluation is performed with the camera mask.

\subsubsection{\MakeUppercase{Image Encoder and Image Size}}
\label{subsubsec:ablation_image}
In Table \ref{tab:image_ablation}, we investigate the effect of varying the image encoder and input image resolution on the mIoU metric, total model parameters, and inference speed (FPS).

\begin{table}[htpb]
\centering
\caption{\textbf{Image encoder and input image size ablation:} A grey row color denotes the choice for the final model. The best-performing model in each category for each input resolution is highlighted in \textbf{bold}.}
\begin{tabular}{cc|ccc}
    \toprule
    \textbf{Encoder} & \textbf{Image Size} & \textbf{mIoU} & \textbf{Model Parameters} & \textbf{FPS} \\
    \midrule
    RN-152 & & \textbf{15.3} & 56.5M & 5.1 \\
    \rowcolor{black!8}
    RN-101 & & 14.9 & 40.9M & 5.4 \\
    RN-50 & 640$\times$384 & 14.8 & 21.9M & 5.6 \\
    RN-34 & & 14.9 & 16.2M & 5.8\\
    RN-18 & & 14.4 & \textbf{10.8M} & \textbf{6.0} \\
    \midrule
    RN-152 & & \textbf{13.8} & 56.5M & 5.4 \\
    RN-101 & & 13.6 & 40.9M & 5.6 \\
    RN-50 & 320$\times$192 & 13.6 & 21.9M & 5.8 \\
    RN-34 & & 13.2 & 16.2M & 6.0 \\
    RN-18 & & 12.8 & \textbf{10.8M} & \textbf{6.2} \\
    \bottomrule
\end{tabular}
\label{tab:image_ablation}
\end{table}

First, using the full input resolution of 640$\times$384, shallower ResNet backbones reduce mIoU and parameter count while improving FPS, as expected. The gap between RN-18 and RN-152 is 0.9 mIoU, suggesting encoder depth is less critical than expected, although a 0.4 gain from RN-101 to RN-152 shows deeper models still provide benefits. While shallower models are typically faster, this is less evident due to the inference bottleneck from grid sampling inherited from OccNeRF and GaussianOcc. As a result, deeper models such as RN-101 or RN-152 may be preferable for a small speed trade-off, though they significantly increase parameters, with RN-152 having 423\% more than RN-18, requiring a balance between performance, speed, and memory.

With reduced input resolution (320$\times$192), performance drops across all models. For example, RN-101 decreases by 1.3 mIoU, highlighting the importance of high-resolution input. A similar trend is observed between RN-18 and RN-152, with a 1 mIoU drop at lower resolution.

\subsubsection{\MakeUppercase{Aggregation of Temporal Samples}}
\label{subsubsec:ablation_samples} 
\begin{figure}[htpb]
    \centering
    \includegraphics[width=.45\textwidth]{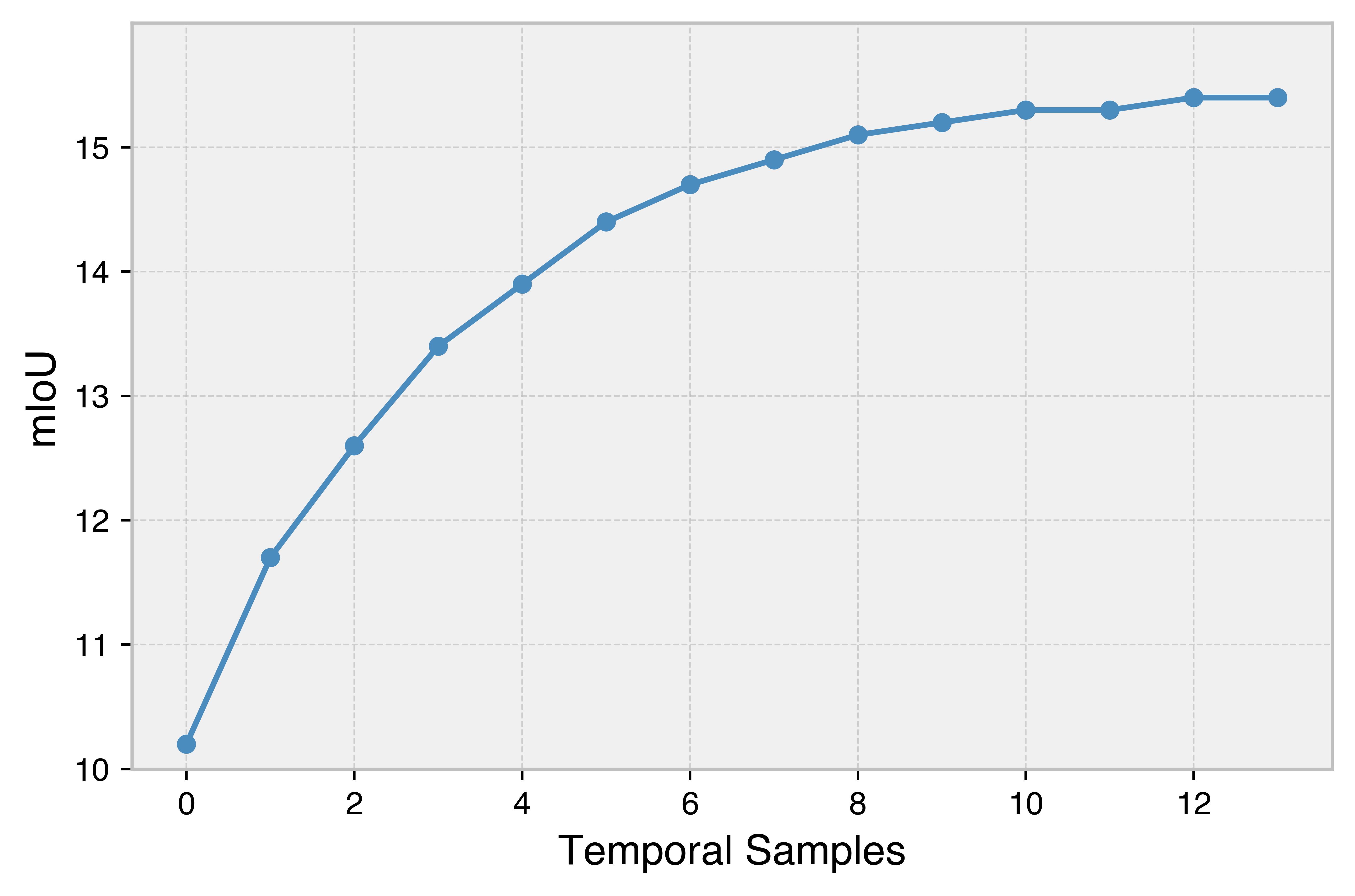}
    \caption{\textbf{Temporal sample ablation.}}
    \label{fig:iou_ablation}
\end{figure}

As shown in the previous Figure \ref{fig:temporal_aggregation}, the use of temporal samples significantly improves the similarity of our 3D pseudo-labels to the Occ3D ground-truth annotations. In Figure \ref{fig:iou_ablation}, we extend this analysis by training the EasyOcc model with varying numbers of temporal samples to evaluate whether a similar trend holds in model performance post-training.

\begin{comment}
    \begin{table}[htpb]
\centering
\begin{tabular}{c|c}
    \hline
    \textbf{Historic Samples} & \textbf{mIoU} \\
    \hline
    0 & 10.08 \\
    1 & 11.23 \\
    2 & 11.99 \\
    3 & 12.29 \\
    5 & 12.77 \\
    7 & 13.14 \\
    10 & \textbf{13.18} \\
    13 & \textbf{13.18} \\
    \hline
\end{tabular}
\vspace{1mm}
\caption{\textbf{Ablation: temporal samples:} Models are trained for 12 epochs to maintain manageable training time. The best-performing configurations are highlighted in \textbf{bold}.}
\label{tab:samples_ablation}
\end{table}
\end{comment}

The figure exhibits a logarithmic curve, seen previously when comparing Occ3D ground-truth labels to the 3D pseudo-labels, with the results clearly showing that increasing the number of temporal samples improves the mIoU metric. This improvement is attributed to scene densification, especially in regions farther from the camera’s field of view. Notably, we achieve an mIoU of 14.9 with both 10 and 13 temporal samples, further indicating a saturation point beyond which additional samples yield diminishing returns in the mIoU metric.

\subsubsection{\MakeUppercase{Variation of $\lambda$}}
\label{subsubsec:ablation_lambda}
In Table \ref{tab:lambda_ablation}, we experiment with the $\lambda$ constant in Equation (\ref{eq:pseudo_loss}) to balance the contributions of the cross-entropy and geometry loss components. The results show that a $\lambda$ value of 0.1 yields the best mIoU performance. A value of 1 performs slightly worse, trailing by 0.2 points, while a value of 0.01 results in a significant drop of 0.7 points. These findings indicate that geometry loss plays a meaningful role, but balancing the contributions of both cross-entropy and geometry losses is crucial. With $\lambda = 0.1$, the magnitudes of the cross-entropy and geometry losses are approximately equal during training, underscoring the importance of careful weighting for optimal performance.

\begin{table}[htpb]
\centering
\caption{\textbf{Pseudo-loss $\lambda$ ablation:} A grey row color denotes the choice for the final model. The best-performing setting is highlighted in \textbf{bold}.}
\begin{tabular}{c|c}
    \toprule
    \textbf{Lambda $\lambda$} & \textbf{mIoU}\\
    \midrule
    0.01 & 14.0 \\
    \rowcolor{black!8}
    0.1 & \textbf{14.9} \\
    1 & 14.7 \\
    \bottomrule
\end{tabular}
\label{tab:lambda_ablation}
\end{table}

\subsubsection{\MakeUppercase{Choice of Losses}}
\label{subsubsec:ablation_losses}
In Table \ref{tab:easyocc_loss_ablation}, we ablate the pseudo-loss in Equation (\ref{eq:pseudo_loss}) by removing individual loss components in Equation (\ref{eq:geometry_loss}), to assess the contribution of each component to the model’s learning.

\begin{table}[htpb]
\centering
\caption{\textbf{Pseudo-loss ablation:} A grey row color denotes the choice for the final model. The best-performing setting is highlighted in \textbf{bold}.}
\begin{tabular}{cccc|c}
    \toprule
    \textbf{Cross En.} & \textbf{Geometry Scale} & \textbf{Semantic Scale} & \textbf{Lovász} & \textbf{mIoU}\\
    \midrule
    \rowcolor{black!8}
    \cmark & \cmark & \cmark & \cmark & 14.9 \\
    \cmark & \cmark & \cmark &  & \textbf{15.1} \\
     & \cmark & \cmark & \cmark & 13.8 \\
     & \cmark & \cmark &  & 14.2 \\
     \cmark &  &  & \cmark & 12.5 \\
     \cmark &  &  &  & 9.2 \\
      &  &  & \cmark & 9.9 \\
    \bottomrule
\end{tabular}
\label{tab:easyocc_loss_ablation}
\end{table}

Overall, the results show that removing components of the loss function reduces the model’s learning capability. For example, excluding the cross-entropy loss decreases mIoU by 1.1 points. Similar drops occur when individual components of the geometry loss are removed. Interestingly, excluding the Lovász-Softmax loss results in a slight mIoU improvement of 0.2, which is unexpected since it is designed to optimize the IoU metric. We hypothesize that this anomaly arises from including the empty class index in the Lovász loss computation. To validate this, we retrain the model excluding this index, as detailed in the following ablation.

\begin{figure*}[ht]
    \centering
    \includegraphics[width=\textwidth]{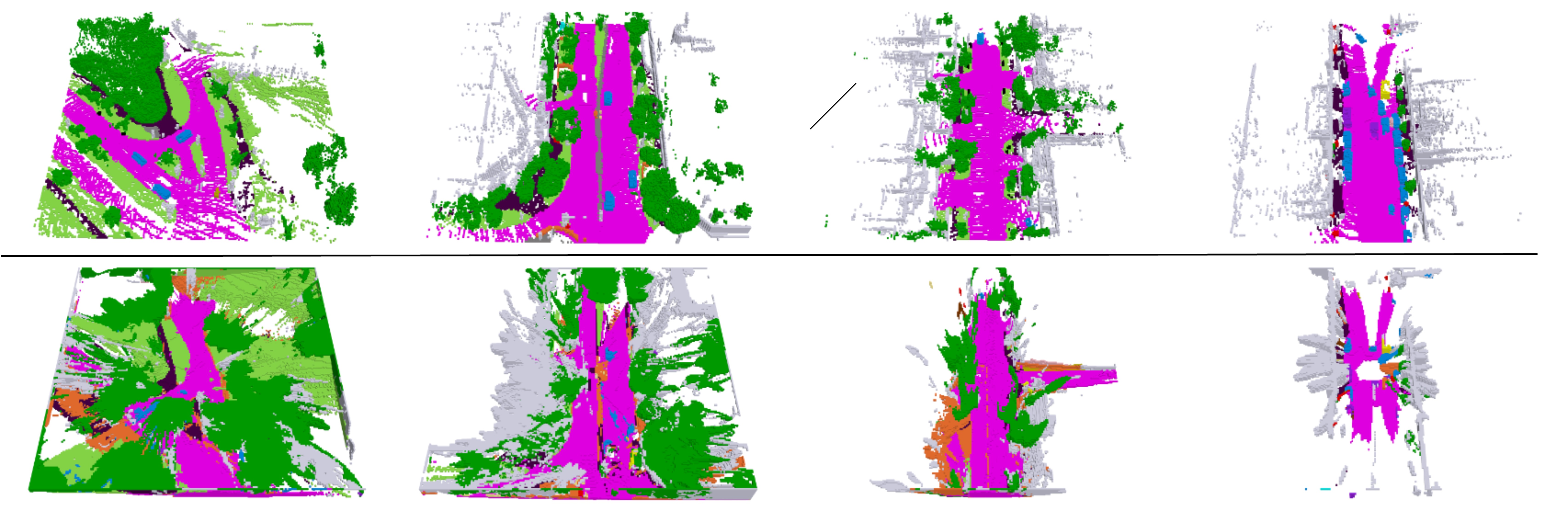}
    \caption{\textbf{Pseudo-label qualitative analysis:} Occ3D-nuScenes \cite{tian2023occ3d} ground truth (\textit{top}) and our Easy3D-Labels (\textit{bottom})}
    \label{fig:pseudo_vis}
\end{figure*}

\subsubsection{\MakeUppercase{Lovàsz Softmax Loss}}
\label{subsubsec:ablation_lovasz}
Following the previous experiment, we retrain the model with the Lovász-Softmax loss that excludes the empty index from the loss computation, as seen in Table \ref{tab:lovasz_ablation}.

\begin{table}[htpb]
\centering
\caption{\textbf{Lovász-Softmax empty index ablation:} A grey row color denotes the choice for the final model. The best-performing setting is highlighted in \textbf{bold}.}
\begin{tabular}{c|c}
    \toprule
    \textbf{Ignore Empty} & \textbf{mIoU}\\
    \midrule
    \xmark & 14.9 \\
    \rowcolor{black!8}
    \cmark & \textbf{15.4} \\
    \bottomrule
\end{tabular}
\label{tab:lovasz_ablation}
\end{table}

Excluding the empty class from the loss computation results in a 0.5 mIoU improvement. Only the previous experiments in the ablation section were conducted with the inclusion of the empty label in the Lovász loss. However, the integrity of the comparative results is maintained, as all models were subject to the same conditions and limitations.

\subsection{\MakeUppercase{Qualitative Results}}
\label{subsec:qual_results}
In this section, we present a qualitative analysis of our Easy3D-Labels versus ground truth (Section \ref{subsec:easy3d_qual}) and occupancy estimation results for the voxel view (Section \ref{subsubsec:qual_voxel}) and from the image view (Section \ref{subsubsec:qual_image}).

\begin{figure*}[ht]
    \centering
    \includegraphics[width=1\textwidth]{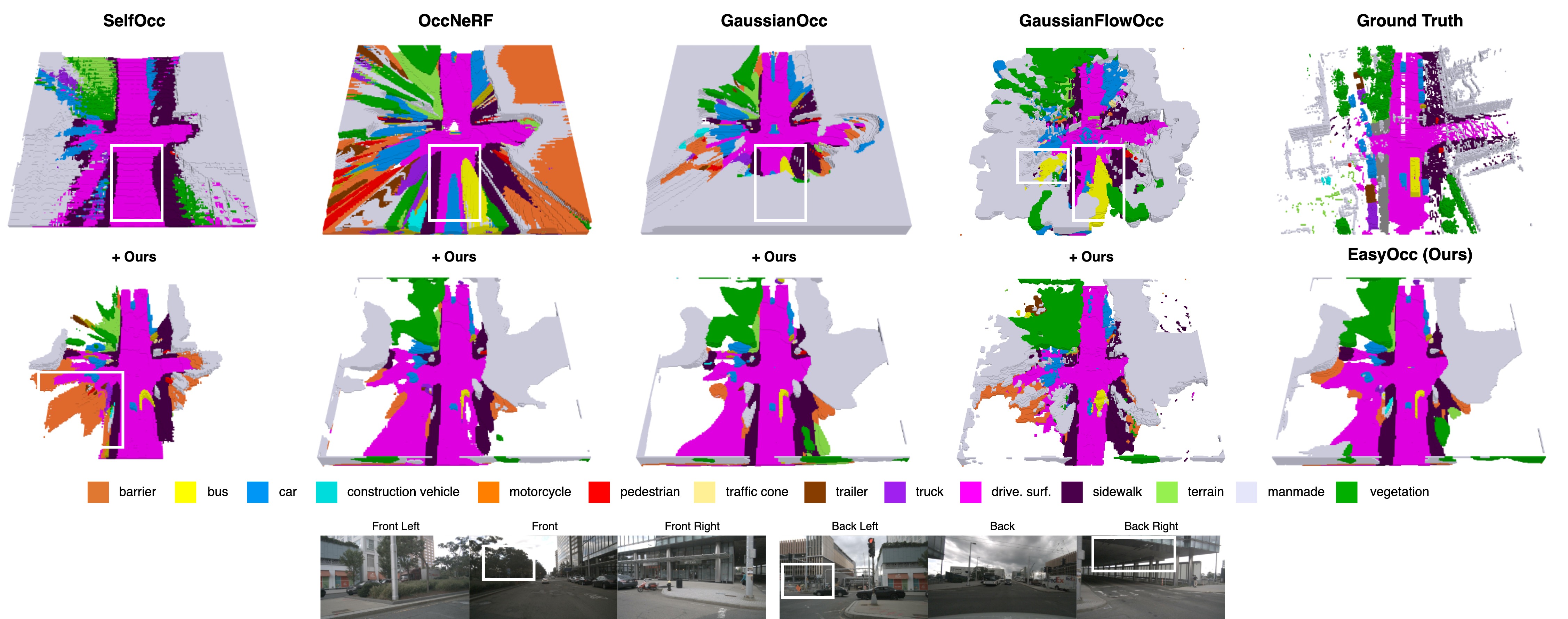}
    \caption{\textbf{Voxel occupancy qualitative analysis} of models evaluated in Table \ref{tab:miou_table}. + Ours indicates the same model trained with Easy3D-Labels.}
    \label{fig:qual_vis1}
\end{figure*}

\begin{figure*}[h]
    \centering
    \includegraphics[width=1\textwidth]{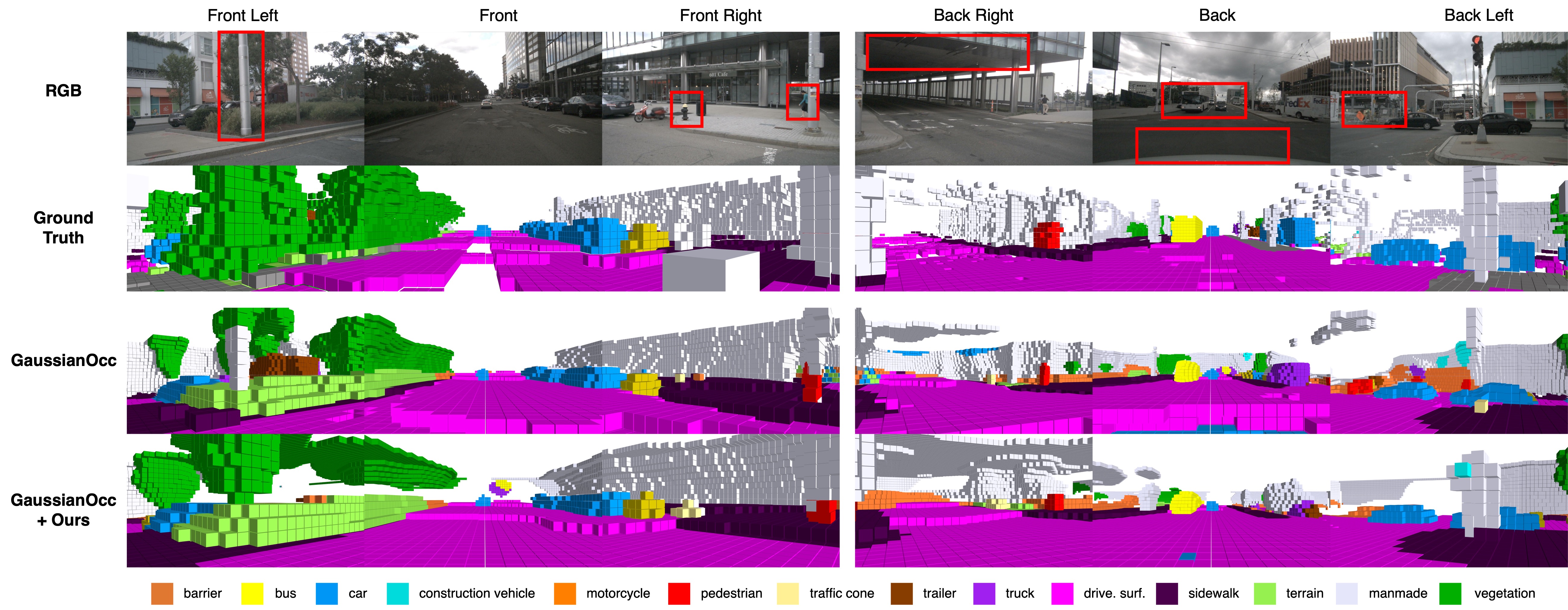}
    \caption{\textbf{Image view occupancy qualitative analysis} of models evaluated in Table \ref{tab:miou_table}. + Ours indicates the same model trained with Easy3D-Labels.}
    \label{fig:qual_vis3}
\end{figure*}

\subsubsection{\MakeUppercase{Easy3D-Label Voxel Analysis}}
\label{subsec:easy3d_qual}
In Figure \ref{fig:pseudo_vis}, we compare four Easy3D-Labels training samples with their corresponding ground-truth labels. The pseudo-labels closely match the ground truth, accurately identifying key scene elements such as roads, vegetation, and buildings. Despite mitigation efforts such as outlier removal and occupancy thresholding, there exist incorrectly labelled voxels primarily due to noise in the depth maps. However, as will be discussed in Section \ref{subsubsec:qual_voxel}, the model's predicted outputs often appear smoother and more continuous than the pseudo-labels. Densification is lacking in the rightmost sample due to the sample being early in the sequence, resulting in limited aggregation of temporal data.

\subsubsection{\MakeUppercase{Voxel Occupancy Estimation}}
\label{subsubsec:qual_voxel}

We begin by analysing Figure \ref{fig:qual_vis1}, which shows semantic voxel visualisations for each model, comparing original models with their pseudo-loss variants.

SelfOcc performs well in road segmentation but misses vehicles in the back camera view; pseudo-loss corrects this but introduces increased misclassification as \textit{barrier}, particularly in the back-left view, likely due to confusing structures such as fences or signs.

OccNeRF predicts more correct objects than SelfOcc, including vehicles in the back view, but suffers from object duplication, leading to penalties in metrics. Pseudo-loss reduces duplication and improves vegetation estimation. Similar improvements are observed in GaussianOcc, where pseudo-loss corrects misclassifications such as the \textit{road} being predicted as a wall.

GaussianFlowOcc suffers from object duplication of the \textit{car} and \textit{bus} classes in the rear view and incorrect detection of a \textit{bus} in the back left view. All of these issues are corrected with the inclusion of Easy3D-Labels.

%Comparing \textit{EasyOcc} and \textit{GaussTR}, GaussTR fails to capture vegetation and building overhangs, while EasyOcc predicts these correctly. However, GaussTR better reconstructs dynamic objects, producing more complete shapes, whereas EasyOcc predicts only visible regions. This aligns with quantitative results, where GaussTR performs better on dynamic classes. EasyOcc, using a voxel representation, produces smoother outputs, while GaussTR’s Gaussian representation appears more fragmented.

\textit{Easy3D-Labels} provide several benefits: they enable correct estimations of regions beneath the ego vehicle through temporal aggregation, improve detection of structures such as building overhangs absent in ground truth, and reduce object duplication and scene densification, leading to improved mIoU and RayIoU. However, due to the label generation process, they mainly predict only visible regions, not correctly filling out areas, such as the entirety of a vehicle.

\subsubsection{\MakeUppercase{Image View Occupancy Estimation}}
\label{subsubsec:qual_image}

Model estimations and our 3D pseudo-labels are generated solely from camera views. We therefore examine semantic voxel estimations of GaussianOcc and its variant trained with Easy3D-Labels across the six camera views in Figure \ref{fig:qual_vis3}.

In the front left view, a ground-truth mislabel causes GaussianOcc to be penalised despite correctly predicting a pole, while our model predicts \textit{vegetation}. In the front right view, both models detect a \textit{pedestrian} but with inaccurate positioning due to occlusion; neither correctly labels smaller objects such as the fire hydrant or trash can, though both detect the \textit{motorcycle}.

In the back right view, our model correctly captures a building overhang, which GaussianOcc misses. In the back view, GaussianOcc fails to predict the \textit{road} in the lower region and incorrectly introduces a wall behind vehicles, while our model produces a continuous \textit{road} and avoids this error. In the back left view, GaussianOcc generates several unsupported estimations (e.g., \textit{barrier}, \textit{pedestrian}, \textit{construction vehicle}), whereas our model produces cleaner and more consistent outputs.

\section{\MakeUppercase{Conclusion}}
\label{sec:conclusion}
This paper presents the use of 3D pseudo-labels, Easy3D-Labels, for self-supervised semantic occupancy estimation models for automated vehicle perception. These labels enable loss computation directly in 3D space, rather than relying on conventional 2D camera-space supervision. They can be easily integrated into existing architectures, leading to improved model performance, more complete scene representation, and better detection of vulnerable road users. We additionally show transferability to models using different scene representations and depth supervision strategies. Moreover, using only these labels for supervision in our model, EasyOcc, proves effective across performance metrics. The strong performance of these 3D pseudo-labels highlights their potential to enhance self-supervised models. However, several directions for future work remain to further refine and evaluate the proposed approach:

\begin{enumerate}
    \item Incorporating LiDAR data into the 3D pseudo-label generation pipeline to facilitate comparison with LiDAR-supervised models.
    \item Conducting a more comprehensive investigation into the integration of 3D pseudo-labels within models that utilise a Gaussian scene representation.
    \item Evaluating the robustness of 3D pseudo-labels under challenging driving conditions, such as rain, fog, and low-light environments.
\end{enumerate}

Self-supervised occupancy estimation methods have historically lagged behind supervised approaches. However, recent advances in 3D supervision, including the results presented in this work, suggest that this performance gap is narrowing. Despite the growing use of sparse scene representations, such as Gaussians, we show that these methods remain compatible with 3D supervision. Our findings indicate that leveraging temporal information and carefully selecting the loss domain are key factors for achieving strong performance in semantic occupancy estimation.

\bibliographystyle{IEEEtran} % Sets the bibliography style
\bibliography{ojvt} % The .bib file name without the extension

\begin{IEEEbiography}[{\includegraphics[width=1in,height=1.25in,clip,keepaspectratio]{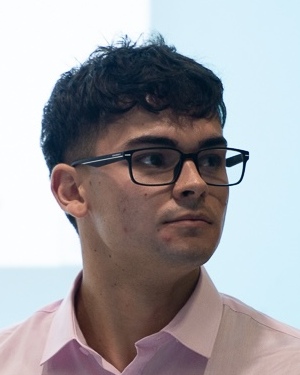}}]{SEAMIE HAYES}~(Graduate Student Member, IEEE) received their B.Sc in Mathematical Sciences from the University of Limerick in 2023. He is pursuing a full-time structured PhD with the CRT Programme in the Department of Electronic and Computer Engineering at the University of Limerick. His area of research is focused on autonomous vehicles with an emphasis on semantic occupancy estimation and scene representation.
\end{IEEEbiography}

\begin{IEEEbiography}[{\includegraphics[width=1in,height=1.25in,clip,keepaspectratio]{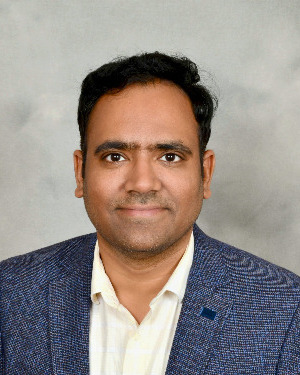}}]{GANESH SISTU}~is a Principal AI Architect at Valeo Ireland and Adjunct Assistant Professor at the University of Limerick, Ireland. He is leading multiple global research teams working on automated driving and parking. With over 14 years in computer vision and machine learning, he has contributed to 35+ publications in top-tier conferences like ICCV and ICRA. He also plays a pivotal role in guiding the future of AI education as an industry board member for Science Foundation Ireland's Data Science PhD and the National MSc in AI programs at the University of Limerick, blending academic insight with industry expertise.
\end{IEEEbiography}

%%%%
%%%%
%%%%
%%%%

\begin{IEEEbiography}[{\includegraphics[width=1in,height=1.25in,clip,keepaspectratio]{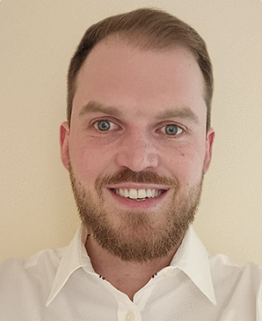}}]{TIM BROPHY}~(Member, IEEE) received the
B.Eng. (Hons.) degree from the University of Galway, Galway, Ireland, in 2018, and the Ph.D. degree with Connaught Automotive Research Group, University of Galway. From 2018 to 2025, he was
a member with Connaught Automotive Research
Group, University of Galway. In 2025, he joined
the University of Limerick, Limerick, Ireland, as
a Postdoctoral Researcher. His research interests
include sensor availability, computer vision, and
artificial intelligence in the context of automated
vehicles.
\end{IEEEbiography}

%%%%
%%%%
%%%%
%%%%

\begin{IEEEbiography}[{\includegraphics[width=1in,height=1.25in,clip,keepaspectratio]{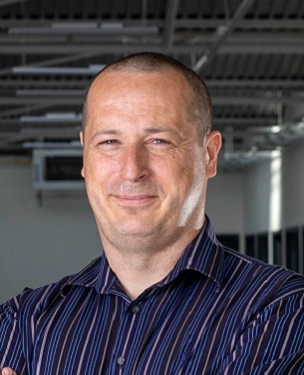}}]{CIARAN EISING}~(Senior Member, IEEE) received
the B.E. degree in electronic and computer engineering and the Ph.D. from NUI Galway, Galway,
Ireland, in 2003 and 2010, respectively. From 2009
to 2020, he was a Computer Vision Architect \&
Senior Expert with Valeo. In 2020, he joined the
University of Limerick, Limerick, Ireland, as an
Associate Professor.
\end{IEEEbiography}

\end{document}